\documentclass[a4paper,fleqn]{cas-dc}

\usepackage[numbers]{natbib}
\usepackage{tabularx}
\usepackage{listings}
\usepackage[flushleft]{threeparttable}
\usepackage{ulem}
 \usepackage[ruled,linesnumbered]{algorithm2e}
 \usepackage{graphicx}
 \usepackage{xcolor}
\definecolor{codered}{rgb}{0.98,0.49,0.72}
\definecolor{codegreen}{rgb}{0,0.75,0}
\def\tsc#1{\csdef{#1}{\textsc{\lowercase{#1}}\xspace}}
\tsc{WGM}
\tsc{QE}
\tsc{EP}
\tsc{PMS}
\tsc{BEC}
\tsc{DE}

\begin{document}
\let\WriteBookmarks\relax
\def\floatpagepagefraction{1}
\def\textpagefraction{.001}
\shorttitle{}
\shortauthors{Wei Luo et~al.}

\title [mode = title]{A Feature Shuffling and Restoration Strategy for Universal Unsupervised Anomaly Detection}                      



\author[1]{Wei Luo}[orcid=0000-0003-3125-054X]
\ead{luow23@mails.tsinghua.edu.cn}


\address[1]{State Key Laboratory of Precision Measurement Technology and Instruments, Tsinghua University, Beijing, 100084, China}

\author[1]{Haiming Yao}[orcid=0000-0003-1419-5489]
\ead{yhm22@mails.tsinghua.edu.cn}

\author[1]{Zhenfeng Qiang}
\ead{18302973462@163.com}

\author[1]{Xiaotian Zhang}
\ead{zhangxt6@foxmail.com}

\author[2]{Weihang Zhang}
\cormark[1]
\ead{zhangweihang@bit.edu.cn}


\address[2]{School of Medical Technology, Beijing Institute of Technology, Beijing, 100081, China}

\cortext[cor1]{Corresponding author}


\begin{abstract}
Unsupervised anomaly detection is vital in industrial fields, with reconstruction-based methods favored for their simplicity and effectiveness. However, reconstruction methods often encounter an identical shortcut issue, where both normal and anomalous regions can be well reconstructed and fail to identify outliers. The severity of this problem increases with  the complexity of the normal data distribution. Consequently, existing methods may exhibit excellent detection performance in a specific scenario, but their performance sharply declines when transferred to another scenario. This paper focuses on establishing a universal model applicable to anomaly detection tasks across different settings,  termed as universal anomaly detection. In this work, we introduce a novel, straightforward yet efficient framework for universal anomaly detection: \uline{F}eature \uline{S}huffling and \uline{R}estoration (FSR), which can alleviate the identical shortcut issue across different settings. First and foremost, FSR employs multi-scale features with rich semantic information as reconstruction targets, rather than raw image pixels. Subsequently, these multi-scale features are partitioned into non-overlapping feature blocks, which are randomly shuffled and then restored to their original state using a restoration network. This simple paradigm encourages the model to focus more on global contextual information. Additionally, we introduce a novel concept, the shuffling rate, to regulate the complexity of the FSR task, thereby alleviating the identical shortcut across different settings. Furthermore, we provide theoretical explanations for the effectiveness of FSR framework from two perspectives: network structure and mutual information. Extensive experimental results validate the superiority and efficiency of the FSR framework across different settings.Code is available at \href{https://github.com/luow23/FSR}{\textcolor{codered}{https://github.com/luow23/FSR}}.

\end{abstract}



\begin{keywords}
Universal anomaly detection\\ Vision transformer\\ Feature reconstruction\\ Feature shuffling and restoration
\end{keywords}

\maketitle
\section{Introduction}
\begin{figure*}
    \centering
    \includegraphics[width=\linewidth]{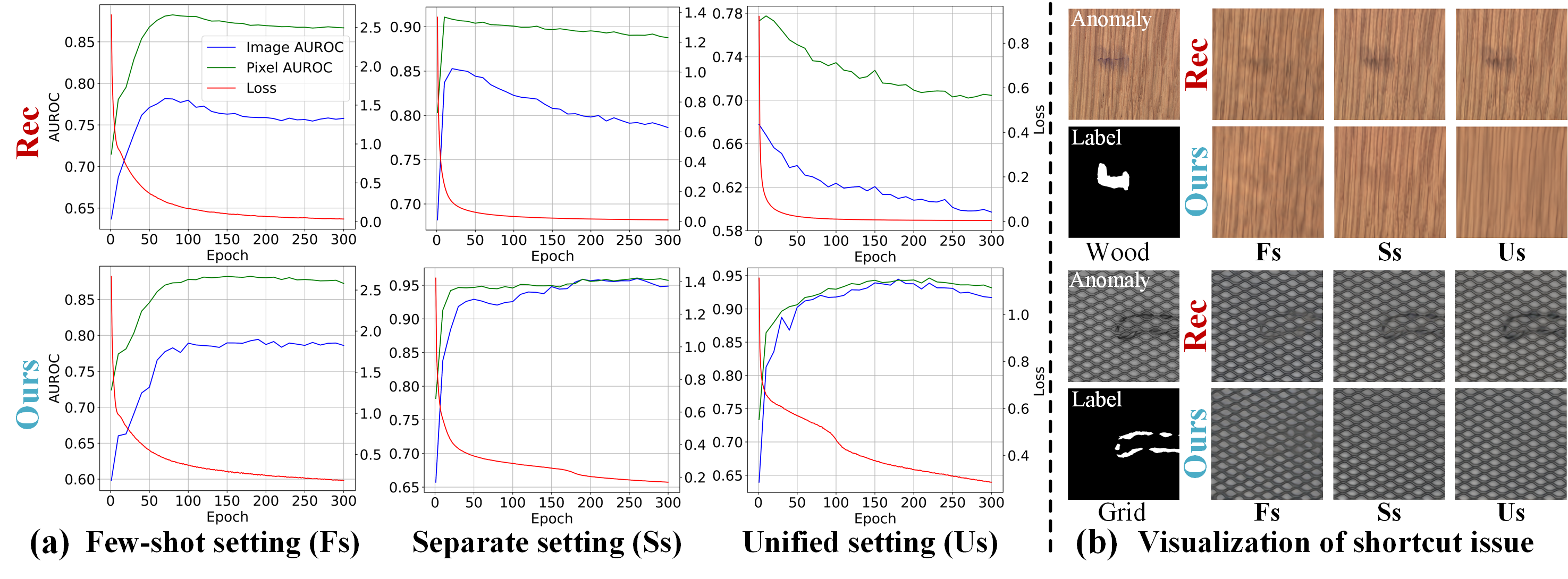}
    \caption{Comparison between reconstruction (Rec) task and our strategy across few-shot, separate, and unified settings. (a) Training loss ({red}) as well as the image/pixel AUROC (\textcolor{blue}{blue}/\textcolor{green}{green}). In the Rec task, transitioning from a few-shot setting (Fs) to a separate setting (Ss) and subsequently to a unified setting (Us), the issue of identical shortcut becomes increasingly pronounced due to the growing complexity of the normal data distribution. On the contrary, our strategy is capable of addressing this issue across different settings. (b) Visualization of the shortcut problem, wherein the anomalous regions can be well reconstructed, making them difficult to distinguish from normal ones. In contrast, our strategy successfully addresses this issue, adeptly reconstructing anomalies as normal samples under various settings. Notably, all models are trained for feature reconstruction. We utilize an additional decoder to visualize the reconstructed features as images. This decoder is only designed for visualization.}
    \label{fig:shortcut}
\end{figure*}

Visual anomaly detection/localization is a binary classification task aimed at determining whether a given image/pixel deviates from the normal pattern, which serves as a crucial component of industrial product control \cite{MVTEC, BTAD, shen2024instrument, luo2024ami}. In real industrial scenarios, the number of abnormal samples is significantly smaller than that of normal samples. This implies that collecting a large number of labeled defective samples for supervised learning \cite{PGANet} is impractical. Therefore, we need to address the anomaly detection task in an unsupervised manner.\\
\indent In unsupervised anomaly detection \cite{ST, wang2025deep, wang2024mtdiff, cao2022informative, bai2024dual, jiang2023masked}, accurately establishing the normal data distribution is crucial. A commonly employed approach to modeling normal data distribution follows the reconstruction \cite{AE,AE-SSIM,DFR,Ganomaly,MemAE} framework. This method posits that a model trained solely on normal data can effectively reconstruct normal patterns during testing but encounters challenges in reconstructing abnormal patterns, leading to larger reconstruction errors for abnormal samples, thereby distinguishing them from normal samples. However, this hypothesis does not always hold, as anomalous patterns can sometimes be perfectly reconstructed, giving rise to the identical shortcut phenomenon. Intuitively, in the context of reconstruction tasks, models tend to favor the direct duplication of input data over the laborious learning of the normal data distribution, thereby overlooking its semantic\begin{figure}
    \centering
    \includegraphics{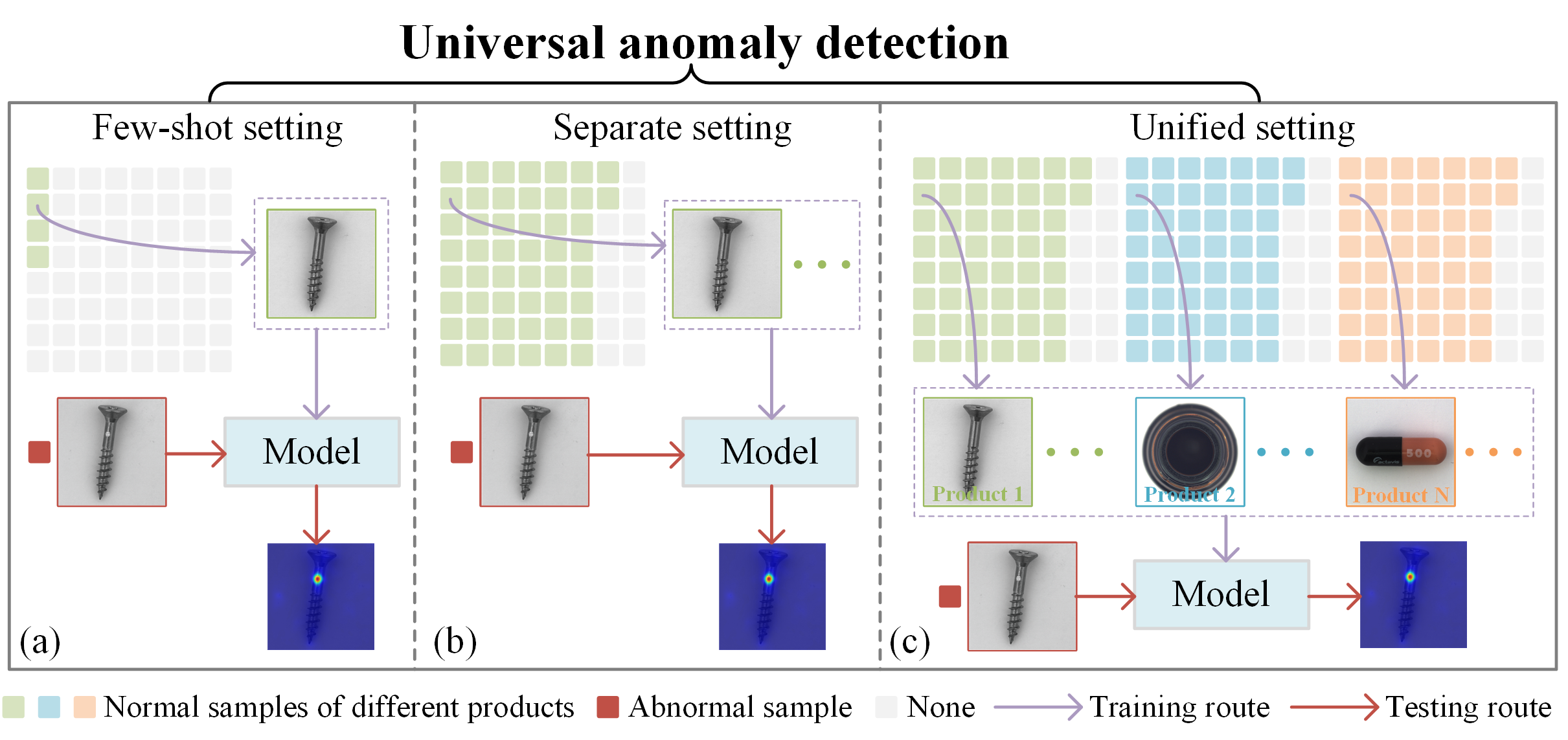}
    \caption{Illustration of universal anomaly detection. The universal anomaly detection refers to comprehensive consideration of a model's detection performance across few-shot, separate, and unified settings. (a) The few-shot setting involves training a model using only a limited number of samples from a specific category of product. (b) In the separate setting, a model is trained using a significant number of samples from a single category of product. (c) The unified setting entails training a single model using a substantial number of samples from multiple categories of products.}
    \label{fig:UAD}
\end{figure} content.\\
\indent Numerous unsupervised anomaly detection methods \cite{RegAD, draem, uniad} have been introduced to tackle the challenge of identical shortcut. However, they overlook a critical aspect: \textbf{in real-world industrial settings, the distribution of normal data undergoes continual changes over time}. In particular, as depicted in Fig. \ref{fig:UAD}, at the initial phase of an industrial production line, only a limited number of normal samples are available, leading to a few-shot setting. As time progresses, a substantial increase in normal samples leads to the establishment of a separate setting. Ultimately, with a diverse array of normal samples from multiple product types, a unified setting emerges. \\
\indent Additionally, we observe that \textbf{the complexity of data distribution directly influences the severity of the identical shortcut issue.} As illustrated in the first row of Fig. \ref{fig:shortcut}(a), progressing from the few-shot \cite{tdg, differnet, RegAD} setting to separate \cite{draem, PacthCore, RD4AD} setting and further to unified \cite{uniad, lu2023hierarchical} setting, the complexity of data distribution gradually increases, and correspondingly, the identical shortcut problem becomes more acute. A more intuitive visualization of this issue is depicted in the first row of Fig. \ref{fig:shortcut}(b).\\
\indent However,   existing methods typically only address this issue for a particular setting, resulting in shortcomings in their transferability. In other words, while these methods may exhibit excellent detection performance in a particular setting, their performance sharply declines when transferred to other settings. For instance, RegAD \cite{RegAD} has demonstrated outstanding detection performance in the few-shot setting by leveraging meta-learning. However, due to its intricate configuration, it faces challenges in transferring to other settings. DRAEM \cite{draem} addresses the issue of identical shortcut in the separate setting by incorporating artificially generated anomalous samples. Nevertheless, its inspection accuracy experiences a significant decline in the unified setting, as the identical shortcut issue becomes more pronounced in this configuration. UniAD \cite{uniad} proposes a layer-wise query decoder and neighbor-masked attention mechanism to alleviate the identical shortcut issue in the unified setting. However, it requires substantial training data to model the normal distribution and is not suitable for few-shot anomaly detection. Therefore, establishing a model capable of exhibiting excellent detection performance across few-shot, separate, and unified settings is highly significant, termed as universal anomaly detection. \\
\indent We believe that the reason for existing methods fail to maintain good performance across various settings is that they address the identical shortcut issue from the perspective of improving the model. {However, we argue that the primary cause of identical shortcut lies in the reconstruction task itself. Because the input and target in reconstruction are identical, the network can easily memorize input features and replicate them without truly understanding the global semantic context, limiting both generalization and anomaly detection performance. To address this issue, we propose a simple yet effective Feature Shuffling and Restoration (FSR) strategy. FSR randomly shuffles a subset of feature blocks and requires the network to restore them to their original positions, thereby forcing the model to capture long-range dependencies and fully leverage global contextual information. This approach effectively mitigates the identical shortcut problem across various settings without necessitating any task-specific modifications to the model.} Firstly, we opt for pre-trained multi-scale features as the reconstruction target, as they possess richer semantic information compared to image pixels. Secondly, for the selection of restoration network, we choose the Vision Transformer (ViT) \cite{Vit}. This is motivated by the fact that FSR task necessitates consideration of the interconnections among individual feature blocks, aligning well with the multi-head attention mechanism inherent in ViT. In contrast, convolutional neural networks (CNNs) exhibit inherent biases such as locality and translational equivariance, rendering them unsuitable for FSR task as they can only account for information within fixed regions of convolutional kernel. Additionally, we introduce a \textbf{shuffling rate} to regulate the difficulty of FSR task, addressing the identical shortcut issue under various settings, as depicted in the second row of Fig. \ref{fig:shortcut}(a). 
{Fig.~\ref{fig:shortcut}(b) clearly illustrates the advantage of FSR: while models trained with conventional reconstruction tasks tend to reproduce anomalous regions, our method effectively restores them.} Finally, we provide theoretical explanations for the effectiveness of FSR from the perspectives of network structure and mutual information. The main contributions of this study
can be summarized as: 
\begin{itemize}
    \item To our knowledge, this is the first attempt to propose a model that exhibits outstanding detection performance across few-shot, separate, and unified settings.
    \item To address the identical shortcut issue, we introduce a simple yet effective feature shuffling and restoration (FSR) strategy, aiding the model in learning global semantics.
    \item We propose a shuffling rate to regulate the difficulty of the FSR proxy task, thereby enabling the model to exhibit optimal performance across various settings.
    \item Without elaborately designed modules, our proposed method achieves outstanding detection performance compared to previous state-of-the-art methods across different settings in the MVTec AD \cite{MVTEC} and BTAD \cite{BTAD} datasets.
\end{itemize}

\section{Related Work}
\subsection{Feature embedding-based methods}
Feature embedding-based methods commonly employ the networks pre-trained on ImageNet dataset \cite{ImageNet} as embedding functions to transform the image space into the compressed feature embedding~\cite{shen2025algorithm, shen2021finger} space. In this embedding space, normal features exhibit close distances to each other, while anomalous features are distantly separated from normal features. Rippel \textit{et al.} \cite{mvgimagelevel} employed a multivariate Gaussian distribution to model image-level features and subsequently used Mahalanobis distance as the criterion for anomaly detection. Building upon this, PaDiM \cite{PaDiM} extends this approach to the pixel level, thus enabling anomaly localization. MBPFM \cite{MBPFM} utilizes two distinct pre-trained networks to extract features, followed by bidirectional mapping of these two sets of features, resulting in superior detection and localization performance. PatchCore \cite{PacthCore} leverages locally aggregated pre-trained features and employs a greedy coreset subsampling algorithm to construct a memory bank of normal features. Subsequently, it uses the distance between test features and the features in the memory bank as the criterion for anomaly detection and localization. Building upon PatchCore, ReconPatch~\cite{hyun2024reconpatch} leverages contrastive learning to enhance the discriminability of features, thereby further boosting detection performance. However, these methods suffer from significant online memory requirements, which limit their applicability in industrial settings.
\subsection{Reconstruction-based methods}
Reconstruction-based methods are based on the assumption that a model trained only on normal samples can reconstruct normal regions but not anomalous ones. Autoencoder (AE) \cite{AE} is a classic method in this category, but it lacks an explicit anomaly suppression mechanism, allowing anomalies to be reconstructed during testing. Therefore, MemAE \cite{MemAE} uses features from the memory bank to replace anomaly features during testing, effectively suppressing the reconstruction of anomalies. PuzzleAE \cite{salehi2020puzzle} transforms the image reconstruction problem into a puzzle-solving task, with the aim of mitigating the issue of overfitting in AE. NDP-Net \cite{NDP-Net} uses a fixed reference normal image to repair defects, resulting in excellent detection performance. FMR-Net \cite{FMR-Net} utilizes memory-generated features to suppress the propagation of defect features in skip connections. DRAEM \cite{draem} leverages natural images to synthesize artificial anomaly defects, thereby transforming the image reconstruction problem into the image restoration problem. DFR \cite{DFR} is a typical feature reconstruction method that utilizes a pre-trained CNN to extract features from images and uses these extracted features as the reconstruction target, thereby enhancing detection performance. The MLDFR \cite{guo2023mldfr} method leverages CNN and ViT to extract local and global pre-trained features separately, subsequently integrating them to obtain a more comprehensive feature representation. The ST-MAE \cite{STMAE} method achieves impressive detection and localization accuracy through a complementary feature transformation strategy. The UTRAD \cite{chen2022utrad} method presents a U-shaped Transformer designed for feature reconstruction, effectively reducing computational complexity while attaining enhanced detection and segmentation accuracy. The FOD \cite{fod} method efficiently inspects defects through patch-wise representation discrepancies and patch-to-normal-patch correlations. Recent methods such as MambaAD~\cite{hemambaad}, Dinomaly~\cite{guo2024dinomaly}, and INP-Former~\cite{Luo_2025_CVPR} have significantly enhanced anomaly detection performance by incorporating advanced architectures, including the State Space Model Mamba~\cite{mamba} and DINO~\cite{DINOV2-R}.\\

\begin{figure*}[!t]
    \centering
    \includegraphics{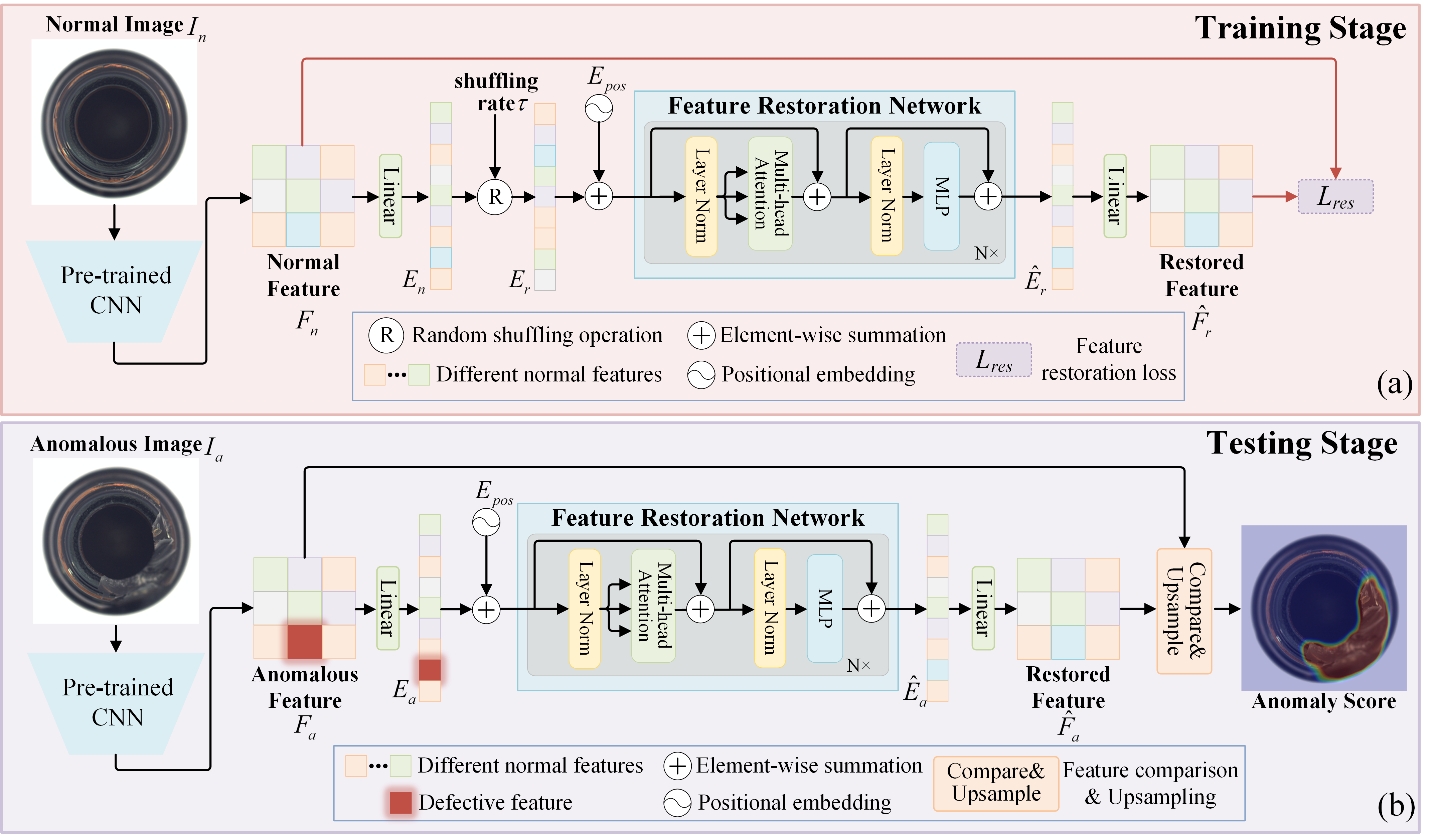}
    \caption{The overall architecture of FSR framework. (a) The training stage comprises three steps: pre-trained feature extraction, random shuffling, and feature restoration. {We use the difference between the input normal feature and  restored feature as the feature restoration loss.} (b) The testing stage incorporates only two steps: pre-trained feature extraction and feature restoration. {We compute the anomaly score by comparing the difference between the input anomalous feature and restored feature, followed by an upsampling operation.}}
    \label{fig:network}
\end{figure*}

\section{Proposed FSR Framework}
\subsection{Model Overview}
This simple yet effective FSR framework is illustrated in Fig. \ref{fig:network}. As shown in Fig. \ref{fig:network}(a), during the training stage, we initially utilize a pre-trained CNN to extract multi-scale features $F_n$ from the normal image $I_n$. Subsequently, these features are randomly shuffled according to the specified shuffling rate $\tau$. The shuffled features then pass through the feature restoration network to obatin reconstructed features $\hat{F}_r$. The disparity between $F_{n}$ and $\hat{F}_r$ constitutes the restoration loss $L_{res}$, which serves as a guidance for the model to learn more meaningful reconstructions. As depicted in Fig. \ref{fig:network}(b), during the testing stage, we directly feed the features $F_a$ of anomalous image $I_a$ into the restoration network to obtain reconstructed features $\hat{F}_a$. Finally, the difference between $F_a$ and $\hat{F}_a$ is utilized for precise defect localization.
\begin{figure}
    \centering
    \includegraphics[width=85mm]{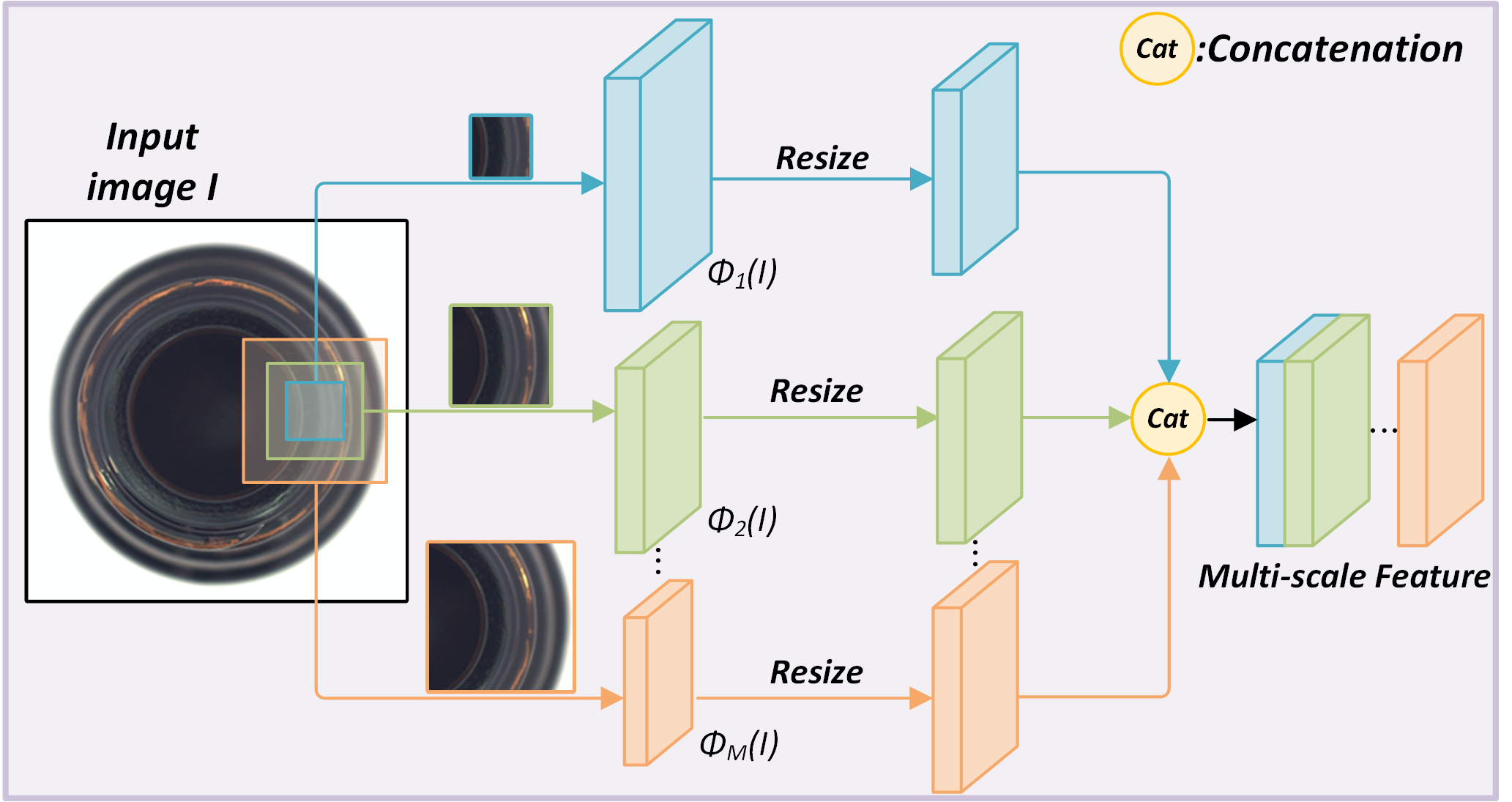}
    \caption{{Illustration of multi-scale feature extraction.}}
    \label{fig:feature}
\end{figure}
\subsection{Pre-trained Feature Extraction}
Salehi \textit{et al.} \cite{MKD} demonstrated that differences between normal and abnormal cases are significantly more pronounced in feature domains than in image domains. Therefore, we employ a CNN, $\phi$, pre-trained on the ImageNet dataset \cite{ImageNet} to transform raw images into feature representations, enhancing the discriminability between normal and abnormal cases.\\
\indent {As shown in Fig.~\ref{fig:feature}, given a normal image $I_n$, we extract feature maps at multiple semantic levels, $\{\phi_1(I_n), \cdots,\phi_M(I_n)\}$, using the pre-trained CNN, where $\phi_m(I_n)\in \mathbb{R}^{H_m\times W_m\times C_m}$. These multi-scale feature maps capture complementary information: shallow layers preserve fine-grained structural details useful for localizing small defects, whereas deeper layers encode high-level semantic context that is crucial for identifying larger or more complex anomalies. Since industrial defects vary significantly in size and shape, we perform cross-scale feature fusion to exploit both fine-grained local cues and global semantic dependencies, resulting in a multi-scale feature map, denoted as $F_n \in \mathbb{R}^{H_F \times W_F \times C_F}$:
\begin{equation}
    F_n= \mho (\Theta(\phi_{1} (I_n)), \cdots, \Theta(\phi_{M} (I_n)))
\end{equation}
where $\Theta$ represents the operation of resizing feature maps to the dimensions $H_F\times W_F$, and $\mho$ signifies the concatenation of feature maps along the channel dimension. The feature maps $F_a$ for the anomalous image $I_a$ are extracted in the same manner as described above.}
\begin{algorithm}[t]
\caption{FSR pseudo-code, PyTorch-like.}
\label{alg:FSR}

\lstset{
  language=Python,
  backgroundcolor=\color{white},
  basicstyle=\fontsize{7.2pt}{7.2pt}\ttfamily\selectfont,
  columns=fullflexible,
  breaklines=true,
  captionpos=b,
  commentstyle=\fontsize{7.2pt}{7.2pt}\color{codegreen},
  keywordstyle=\fontsize{7.2pt}{7.2pt}\color{codered},
}

\begin{lstlisting}[language=python]
# model_fs: network for feature restoration
# x: input feature
# s: shuffling rate
# x_s: shuffled feature
# rec_x: reconstructed feature
# rec_loss: reconstruction loss in Eq. (13)

x_s = random_shuffle(x, s) # feature shuffling
rec_x = model_fs(x_s) # feature restoration
loss = rec_loss(x, rec_x) # compute loss
loss.backward() # back-propagate
update(model_fs) # AdamW update

def random_shuffle(x, s):
    # N:batchsize L:sequence length D:embedding dimension
    N, L, D = x.shape
     # calculate the number of features to shuffle
    num_s = int(L * s)
    # generate random indices for shuffling
    index_s = torch.randperm(L)[:num_s]
    # create a copy of the original features
    x_s = x.clone()
    # shuffle selected features by swapping positions
    index_r = index_s[random.sample(range(num_s), num_s)]
    x_s[:,index_s, :] = x[:, index_r, :]
    return x_s
\end{lstlisting}
\end{algorithm}
\subsection{Feature Shuffling and Restoration}
Reconstruction-based methods suffer from the shortcut learning problem, where anomalies are perfectly reconstructed during testing. To address this issue, we propose a simple yet effective task: feature shuffling and restoration.\\
\indent As shown in Fig. \ref{fig:network}(a), after obtaining multi-scale feature maps $F_n$, our first step is to partition them into non-overlapping feature blocks $E_n$ of size $P$. Following the ViT \cite{Vit} approach, we achieve this using a linear projection with a convolutional kernel of size $P$.
\begin{equation}
    E_n = f^1_{linear}(F_n; \theta^1_{linear})
\end{equation}
where $E_n = \{E^1_n, E^2_n, \cdots, E^L_n|E^l_n\in \mathbb{R}^D\}$, $L=\frac{H_F}{P}\times \frac{W_F}{P}$, $f^1_{linear}$ and $\theta^1_{linear}$ represent the function and parameters of the linear projection, and $D$ denotes the number of channels.\\
\indent Next, we perform random shuffling operation on $E_n$, such as reshuffling $\{E^1_n, E^2_n, \cdots, E^L_n\}$ into $\{E^L_n, E^1_n, \cdots, E^2_n\}$. In deep learning, a noteworthy issue is the alignment between model capacity and agent task difficulty. When the agent task difficulty significantly exceeds the model's capacity, it can lead to difficulties in model fitting. Conversely, when the model's capacity far surpasses the agent task difficulty, it can result in model overfitting. When we randomly shuffle all feature blocks in $E_n$, it leads to the former scenario, whereas performing feature reconstruction without shuffling leads to the latter scenario. Hence, we introduce a shuffling rate $\tau$ to control the training task's difficulty, thus enabling the exploration of the model's optimal performance. In $E_n$, there are a total of $L$ feature blocks. We randomly select $L\times \tau$ feature blocks for shuffling, while keeping the remaining $L\times (1-\tau)$ feature blocks unchanged. This process yields the shuffled feature sequence $E_r$.{
\begin{equation}
\begin{aligned}
    & E_r = R(E_n, \tau) + E_{pos}, \\
    & E_{pos}(l,2i) = \sin\!\left(\frac{l}{10000^{2i/D}}\right), \\
    & E_{pos}(l,2i+1) = \cos\!\left(\frac{l}{10000^{2i/D}}\right),
\end{aligned}
\end{equation}
where $R$ represents the random shuffling operation. Here, $E_{pos} \in \mathbb{R}^{L \times D}$ denotes the fixed sinusoidal positional encoding, where $l \in [0, L-1]$ is the position index of the feature block, $i$ indexes the embedding dimension, and $D$ is the feature dimension.\\
\indent The positional encoding injects location information into the feature sequence, enabling the model to distinguish the order and spatial context of feature blocks even after shuffling. Importantly, $E_{pos}$ is computed solely based on the original order of $E_n$ and remains unaltered during shuffling, thereby preserving the notion of absolute position. Adding $E_{pos}$ to $R(E_n, \tau)$ is the final step in the shuffling procedure, ensuring that the restoration network can leverage both the shuffled content and the original positional information to recover spatially coherent features. We adopt sinusoidal positional encoding rather than learnable embeddings, as it introduces no additional parameters and generalizes naturally to arbitrary sequence lengths.}\\
\indent Subsequently, we feed $E_r$ into the restoration network $f_{res}$ composed of $N$ vision transformer blocks, resulting in the restored feature sequence $\hat{E}_r$.
\begin{equation}
    \hat{E}_r = f_{res}(E_r;\theta_{res})
\end{equation}
where $f_{res}$ and $\theta_{res}$ respectively represent the function and parameters of the restoration network. Subsequently, we utilize another linear projection $f^2_{linear}$ to transform the restored feature sequence $\hat{E}_r$ into the feature map $\hat{F}_r$.
\begin{equation}
    \hat{F}_r = f^2_{linear}(\hat{E}_r;\theta^2_{linear})
\end{equation}
where $\hat{F}_r\in \mathbb{R}^{H_F\times W_F\times C_F}$, $f^2_{linear}$ and $\theta^2_{linear}$ respectively denote the function and parameters of the linear projection.\\
\indent Algorithm \ref{alg:FSR} summarizes the training procedure, together with the pseudo-code of feature shuffling and restoration.
\begin{figure}[!t]
    \centering
    \includegraphics[width=85mm]{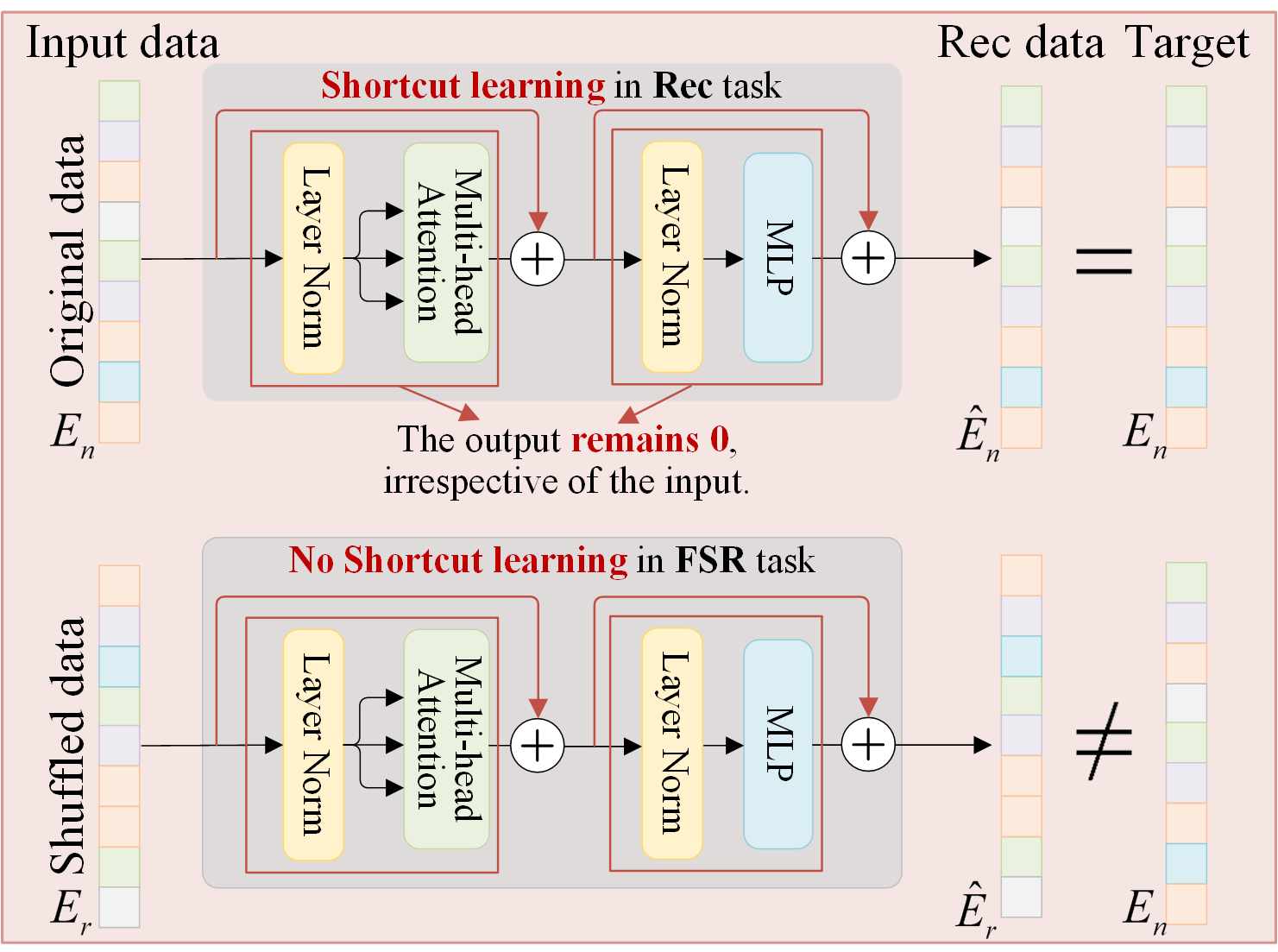}
    \caption{The intuitive explanation of feature shuffling and restoration (FSR) effectiveness from the perspective of network structure. In reconstruction (Rec) task, shortcut learning issues exist, while there are no shortcut learning issues in FSR task.} 
    \label{fig:explanation-network}
\end{figure}
\subsection{Theoretical Explanation}
In this subsection, we elaborate on the superiority of  feature shuffling and restoration (FSR) task compared with feature reconstruction (Rec) task from two aspects: network structure and mutual information.
\subsubsection{Network structure}
\label{explanation-ns}
As illustrated in Fig. \ref{fig:explanation-network}, we provide an intuitive explanation of the superiority of FSR task over Rec task from the pespective of the ViT structure. For the purpose of simplified analysis, we consider a single layer transformer block, which includes LayerNorm (LN), Multi-head Self-Attention (MSA), and a Multi-Layer Perceptron (MLP). In Fig. \ref{fig:explanation-network}(a), we first consider the Rec task, where the input is the original data $E_n$, and the reconstructed $\hat{E}_n$ is obtained using ViT.
\begin{equation}
\label{formal:vit}
\begin{aligned}
    E^{'}_n=E_n+MSA(LN(E_n))\\
    \hat{E}_n=E^{'}_n+MLP(LN(E^{'}_n))
\end{aligned}
\end{equation}
The objective of the Rec task is to make the reconstructed data $\hat{E}_n$ as close as possible to the target $E_n$. However, from the first row of Fig. \ref{fig:explanation-network} and Eq. \ref{formal:vit}, we observe a clear case of shortcut learning in the Rec task. In this scenario, regardless of the input data, setting both $MSA(LN(E_n))$ and $MLP(LN(E^{'}_n))$ to output \textbf{0} would minimize the reconstruction error, leading to a perfect reconstruction of anomalies during the testing phase. In contrast, there is no such shortcut learning issue in the FSR task. The original $E_n$ goes through random shuffling to obtain shuffled data $E_r$, which is then input into ViT to produce the reconstructed data $\hat{E}_r$. As shown in the second row of Fig. \ref{fig:explanation-network}, if both $MSA(LN(E_r))$ and $MLP(LN(E^{'}_r))$ output \textbf{0}, the reconstructed data $\hat{E}_r$ would be the same as the shuffled data $E_r$ and entirely different from the target $E_n$. This would lead to effective reconstruction loss and weight updating, thus effectively circumventing such shortcut learning issues. 
\begin{figure}[!t]
    \centering
    \includegraphics[width=85mm]{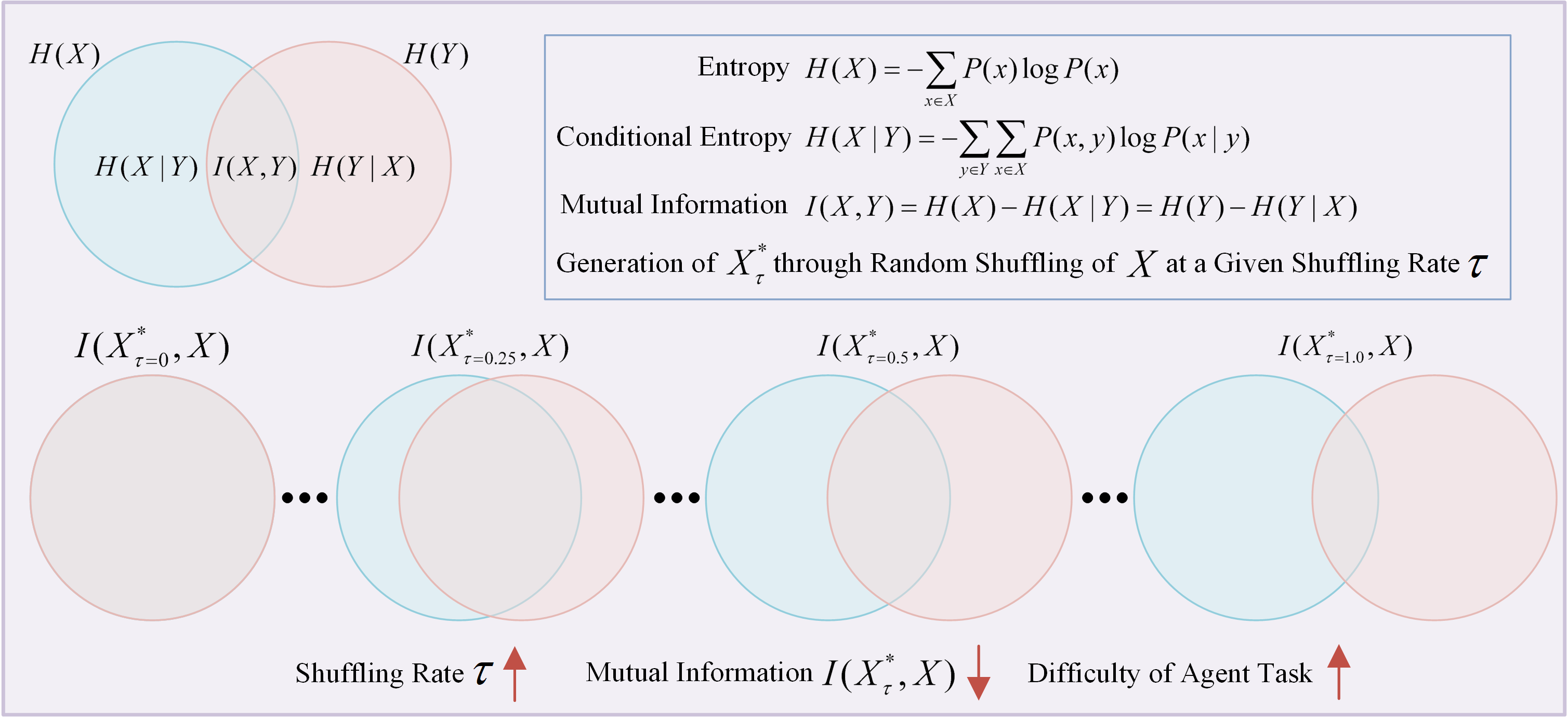}
    \caption{The mathematical explanation of FSR effectiveness from the perspective of mutual informaton.}
    \label{fig:explanation-mutualinformation}
\end{figure}
\subsubsection{Mutual information}
As shown in Fig. \ref{fig:explanation-mutualinformation}, we provide a mathematical explanation of FSR effectiveness from the perspective of mutual information. To begin with, we introduce some fundamental concepts in information theory: entropy, conditional entropy, and mutual information. 
\begin{gather}
 H(X)=-\sum_{x\in X}P(x)logP(x) \notag\\
H(X|Y)=-\sum_{y\in Y}\sum_{x\in X}P(x,y)logP(x|y)\\
I(X,Y)=H(X)-H(X|Y)=H(Y)-H(Y|X) \notag
\end{gather}
where $X$ and $Y$ are random variables, $P(\cdot)$ denotes the probability, $H(X)$ represents the entropy of $X$, $H(X|Y)$ signifies the conditional entropy of $X$ with respect to $Y$, and $I(X, Y)$ denotes the mutual information between $X$ and $Y$. Next, we turn our attention to the FSR task. Here, we define $X$ as a random variable representing the input original features, and $X^*_{\tau}$ represents the features obtained by shuffling the original features $X$ at a given shuffling rate $\tau$ ($0\le \tau \le 1$, when $\tau=0$, the FSR task is essentially the Rec task). $X$ serves as the reconstruction target. We provide the mutual information between input shuffled data $X^{*}_{\tau}$ and target $X$.
\begin{align}
     I(X^*_{\tau},X)&=H(X)-H(X|X^*_{\tau}) \notag\\
               &=H(X)+\sum_{x^*_{\tau}\in X^*_{\tau}}\sum_{x\in X}P(x,x^*_{\tau})logP(x|x^*_{\tau})\\
               &=H(X)+E(logP(x|x^*_{\tau})) \notag
\end{align}
where $E(\cdot)$ represents the expectation. As the shuffling rate $\tau$ increases from 0 to 1, the first term $H(X)$ in the above formula remains constant, while the second term $E(logP(x|x^*_{\tau}))$ steadily decreases. Consequently, as shown in the second row of Fig. \ref{fig:explanation-mutualinformation}, this leads to a continuous reduction in the mutual information between $X^{*}_{\tau}$ and $X$, making the agent task more challenging.
\begin{equation}
\underbrace{I(X^*_{\tau=0},X)}_{Rec\ task}>\underbrace{ \cdots>I(X^*_{\tau=0.5},X)>\cdots>I(X^*_{\tau=1.0},X)}_{FSR\ task}
\end{equation}
\indent Therefore, from the perspective of mutual information, the Rec agent task becomes too straightforward, leading the model to not consider semantic information. Consequently, the model can achieve the agent task by directly copying input data, without accurately modeling the distribution of normal data. However, when $\tau=1.0$ in the FSR task, it becomes excessively challenging, making it difficult for the model to perform the agent task effectively. Hence, determining an appropriate shuffling rate is crucial, as it adjusts the difficulty of the FSR task, allowing the network to accurately model the normal data distribution under a fixed setting. The ablation experiment concerning $\tau$ in Section \ref{effectshuffling} substantiate our hypothesis.

\begin{table*}[!h]
\centering
\caption{Anomaly detection and localization results in terms of image/pixel level AUROC on MVTec AD dataset \cite{MVTEC} under the \textbf{separate} setting.} 
\label{table:separate}
\fontsize{10}{14}\selectfont{
\resizebox{\textwidth}{!}{
\begin{tabular}{cc|ccccccccc|>{\columncolor{orange!15}}cc}
\toprule[1.0pt]
\multicolumn{2}{c|}{\multirow{2}{*}{Category}}             & PuzzleAE & TrustMAE  & RIAD      & DFR       & CutPaste  & DRAEM     & RD4AD     & MBPFM     & PatchCore & \textbf{Ours} & \textbf{Ours}  \\ 
\multicolumn{2}{c|}{}                                      & \cite{salehi2020puzzle}         &\cite{TrustMAE}           &\cite{RIAD}           &\cite{DFR}           &\cite{li2021cutpaste}           &\cite{draem}           &\cite{RD4AD}          &\cite{MBPFM}           &\cite{PacthCore}          &($\tau=0.1$) &($\tau=0.3$)              \\ \midrule
\multicolumn{1}{c|}{\multirow{5}{*}{\rotatebox{90}{Texture}}}                  & Carpet     & 65.7/-   & 97.4/98.5 & 84.2/94.2 & -/97.0    & 93.9/98.3 & 97.0/95.5 & 98.9/98.9 & \textbf{100}/\textbf{99.2}  & 98.7/\uline{99.0} & \uline{99.3}/\textbf{99.2} & \textbf{100}/\uline{99.0}             \\
\multicolumn{1}{c|}{}                         & Grid       & 75.4/-   & 99.1/97.5 & 99.6/96.3 & -/98.0    & \textbf{100}/97.5  & \uline{99.9}/\textbf{99.7} & \textbf{100}/\uline{99.3}  & 98.0/98.8 & 98.2/98.7 & \uline{99.9}/98.9 & \uline{99.9}/99.0             \\
\multicolumn{1}{c|}{}                         & Leather    & 72.9/-   & 95.1/98.1 & \textbf{100}/\uline{99.4}  & -/98.0    & \textbf{100}/\textbf{99.5}  & \textbf{100}/98.6  & \textbf{100}/\uline{99.4}  & \textbf{100}/\uline{99.4}  & \textbf{100}/98.3  & \textbf{100}/99.3  & \textbf{100}/\uline{99.4}             \\
\multicolumn{1}{c|}{}                         & Tile       & 65.5/-   & 97.3/82.5 & 98.7/89.1 & -/87.0    & 94.6/90.5 & \uline{99.6}/\textbf{99.2} & 99.3/95.6 & \uline{99.6}/\uline{96.2} & 98.7/95.6 & \textbf{100}/96.0 & \textbf{100}/95.2        \\
\multicolumn{1}{c|}{}                         & Wood       & 89.5/-   & \textbf{99.8}/92.6 & 93.0/85.8 & -/94.0    & 99.1/95.5 & 99.1/\textbf{96.4} & 99.2/95.3 & \uline{99.5}/\uline{95.6} & 99.2/95.0 & \uline{99.5}/95.1 & \uline{99.5}/95.3             \\ \midrule
\multicolumn{1}{c|}{\multirow{10}{*}{\rotatebox{90}{Object}}} & Bottle     & 94.2/-   & 97.0/93.4 & 99.9/98.4 & -/97.0    & 98.2/97.6 & 99.2/\textbf{99.1} & \textbf{100}/\uline{98.7}  & \textbf{100}/98.4  & \textbf{100}/98.6  & \textbf{100}/\uline{98.7} & \textbf{100}/\uline{98.7}       \\
\multicolumn{1}{c|}{}                         & Cable      & 97.9/-   & 85.1/92.9 & 81.9/84.2 & -/92.0    & 81.2/90.0 & 91.8/94.7 & 95.0/97.4 & 98.8/96.7 & \uline{99.5}/98.4 &  \textbf{99.9}/\textbf{99.0}& \textbf{99.9}/\uline{98.9}             \\
\multicolumn{1}{c|}{}                         & Capsule    & 66.9/-   & 78.8/87.4 & 88.4/92.8 & -/99.0    & 98.2/97.4 & \textbf{98.5}/94.3 & 96.3/98.7 & 94.5/98.3 & 98.1/\uline{98.8} & 97.6/\textbf{99.0} & \uline{98.3}/\textbf{99.0}             \\
\multicolumn{1}{c|}{}                         & Hazelnut   & 91.2/-   & 98.5/98.5 & 83.3/96.1 & -/99.0    & 98.3/97.3 & \textbf{100}/\textbf{99.7}  & \uline{99.9}/98.9 & \textbf{100}/\uline{99.1}  & \textbf{100}/98.7  & \textbf{100}/98.9 & \textbf{100}/98.9              \\
\multicolumn{1}{c|}{}                         & Metal Nut  & 66.3/-   & 76.1/91.8 & 88.5/92.5 & -/93.0    & \uline{99.9}/93.1 & 98.7/\textbf{99.5} & \textbf{100}/97.3  & \textbf{100}/97.2  & \textbf{100}/\uline{98.4}  & 99.5/97.4  & 99.7/97.3             \\
\multicolumn{1}{c|}{}                         & Pill       & 71.6/-   & 83.3/89.9 & 83.8/95.7 & -/97.0    & 94.9/95.7 & \textbf{98.9}/97.6 & 96.6/\uline{98.2} & 96.5/97.2 & 96.6/97.4 & \uline{97.0}/\textbf{98.8}& \uline{97.0}/\textbf{98.8}             \\
\multicolumn{1}{c|}{}                         & Screw      & 57.8/-   & 83.4/97.6 & 84.5/98.8 & -/99.0    & 88.7/96.7 & 93.9/97.6 & 97.0/99.6 & 91.8/98.7 & \uline{98.1}/99.4 & \uline{98.1}/\uline{99.5} & \textbf{98.6}/\textbf{99.6}             \\
\multicolumn{1}{c|}{}                         & Toothbrush & 97.8/-   & 96.9/98.1 & \textbf{100}/\uline{98.9}  & -/98.1    & 99.4/98.1 & \textbf{100}/98.1  & \uline{99.5}/\textbf{99.1} & 88.6/98.6 & \textbf{100}/98.7  &97.2/\uline{98.9} & 97.2/\uline{98.9}             \\
\multicolumn{1}{c|}{}                         & Transistor & 86.0/-   & 87.5/92.7 & 90.9/87.7 & -/80.0    & 96.1/93.0 & 93.1/90.9 & 96.7/92.5 & \uline{97.8}/87.8 & \textbf{100}/96.3  & \textbf{100}/\uline{99.0} & \textbf{100}/\textbf{99.1}              \\
\multicolumn{1}{c|}{}                         & Zipper     & 77.7/-   & 87.5/97.8 & 98.1/97.8 & -/96.0    & \uline{99.9}/\textbf{99.3} & \textbf{100}/\uline{98.8}  & 98.5/98.2 & 97.4/98.2 & 99.4/98.5 & 98.5/98.5  & 98.5/98.6             \\ \midrule
\multicolumn{2}{c|}{Mean}                                  & 77.6/-   & 90.9/94.0 & 91.7/94.2 & 93.8/95.5 & 96.1/96.0 & 98.0/97.3 & 98.5/97.8 & 97.5/97.3 & \uline{99.1}/\uline{98.1} & \uline{99.1}/\textbf{98.4} & \textbf{99.2}/\textbf{98.4}             \\
\bottomrule[1.0pt]
\end{tabular}}}
\end{table*}
\subsection{Training and Testing Procedure}
\subsubsection{Training procedure}
In most of the reconstruction methods \cite{DFR, salehi2020puzzle}, they simply employ local Mean Squared Error (MSE) loss and cosine similarity as the loss functions, without considering global information, while global information contributes to the reconstruction of higher-quality features. Therefore, we use both local and global loss functions to jointly optimize the FSR model. It is worth noting that our local loss function is composed of both MSE loss and cosine similarity. Given a normal feature map $F_n\in \mathbb{R}^{H_F\times W_F \times C_F}$, FSR randomly shuffles and feeds it into the restoration network to obtain the reconstructed feature map $\hat{F}_r\in \mathbb{R}^{H_F\times W_F \times C_F}$. The restoration loss function $L_{res}$ between them is defined as follows:
\begin{equation}
    L_{local}^{mse}=\frac{1}{H_FW_F}\sum_{h=1}^{H_F}\sum_{w=1}^{W_F}{|| F_n(h,w)-\hat{F}_r(h,w) || }_2^2
\end{equation}
\begin{equation}
    L_{local}^{cos}=\frac{1}{H_FW_F}\sum_{h=1}^{H_F}\sum_{w=1}^{W_F}1-\frac{F_n(h,w)^T\cdot \hat{F}_r(h,w)}{|| F_n(h,w) ||\,|| \hat{F}_r(h,w) ||   }
\end{equation}
\begin{equation}
    L_{global}=1-\frac{vec(F_n)^T\cdot vec(\hat{F}_r )}{||vec(F_n)||\,||vec(\hat{F}_r)|| } 
\end{equation}
\begin{equation}
    L_{res} = \underbrace{L_{local}^{mse}+L_{local}^{cos}}_{local}+\underbrace{L_{global}}_{global}
\end{equation}
where ${||\cdot||}_2^2$, $\cdot$, and $||\cdot||$ denote the $L_2$ norm, inner product, and modulus length, respectively. The operation $vec(\cdot)$ represents a flattening process that transforms a 2-D feature map $F \in \mathbb{R}^{H_F\times W_F \times C_F}$ into a vector $v \in \mathbb{R}^{H_FW_FC_F}$. 
\subsubsection{Testing procedure}
During the testing phase, FSR can accurately detect and localize anomalies. Given an anomaly feature map $F_a\in \mathbb{R}^{H_F\times W_F \times C_F}$, FSR directly feeds it into the restoration network to obtain the reconstructed feature map $\hat{F}_a\in \mathbb{R}^{H_F\times W_F \times C_F}$. The discrepancy between $F_a$ and $\hat{F}_a$ can be utilized for anomaly localization. Consistent with the training process, we employ two metrics, MSE and cosine similarity, to quantify the disparity between them.
\begin{equation}
    AS_{mse}(h,w)={||F_a(h,w)-\hat{F}_a(h,w)||}_2^2
\end{equation}
\begin{equation}
    AS_{cos}(h,w)=1-\frac{F_a(h,w)^T\cdot \hat{F}_a(h,w)}{|| F_a(h,w) ||\,|| \hat{F}_a(h,w) ||}
\end{equation}
\begin{equation}
    AS_{final}=Resize(AS_{mse}\otimes AS_{cos})
\end{equation}
where $AS_{mse},AS_{cos}\in \mathbb{R}^{H_F\times W_F}$, $\otimes$ denotes the element-wise product, and $Resize(\cdot)$ denotes the operation of resizing to the original image size. The standard deviation of $AS_{final}$ is used for image-level anomaly detection.

\section{Experiments}
\subsection{Experimental Configuration}
\subsubsection{Dataset}
\textbf{MVTec AD} \cite{MVTEC} is a widely used anomaly detection dataset consisting of 15 categories of industrial defect images, comprising 5 texture and 10 object categories. It comprises 3629 normal images for training, along with 498 normal and 1982 defect images for testing. The \textbf{BTAD} \cite{BTAD} dataset comprises three industrial products: Product 01, Product 02, and Product 03. It consists of 1799 defect-free samples for training, along with 451 defect-free samples and 290 anomalous samples for testing. The intricate texture complexity presents a significant challenge for anomaly detection.
\subsubsection{Implementation Details}
The input image size of MVTec AD is 256 $\times$ 256 $\times$ 3, and the size for resizing feature maps is set as 64 $\times$ 64. The feature maps from 2nd to 4th layers of WideResNet50 \cite{wideresnet} are resized and concatenated together to form a 1792-channel feature map. The default patch size $P$ is 4. The restoration network consists of 8 transformer blocks with hidden dimension of 768 and 12 attention heads. \textbf{{To balance the model's performance across different settings, we set the default shuffling rate to 0.1.}} {\textbf{The optimal shuffling rates $\tau$ for few-shot, separate, and unified settings are 0.1, 0.3, and 0.9, respectively}}. The FSR is trained for 300 epochs using AdamW \cite{AdamW} optimizer with learning rate $1 \times 10^{-3}$ and weight decay $1 \times 10^{-4}$. The batch size is set to 1, 8, and 8 in the few-shot, separate, and unified settings, respectively. All experiments were performed on a computer equipped with an Intel(R) Xeon(R) Silver 4116 CPU running at 2.10 GHz and an NVIDIA Tesla A100 GPU. {Unless otherwise specified, a fixed random seed of 1 was used for all experiments to ensure reproducibility.}
\subsubsection{Evaluation Metrics}
The widely used Area Under the Receiver Operating Characteristic Curve (AUROC) metrics at both the image and pixel levels serve as the evaluation standard for anomaly detection and localization.
\subsubsection{{Symbol Definitions}}
{For clarity, the symbols used in the following tables are defined as follows: the best result is shown in \textbf{bold}, and the second-best result is \underline{underlined}. $\dag$ represents the results obtained through our reproduction, and our default setting is marked in} {\setlength{\fboxsep}{2pt}\colorbox{orange!15}{{orange}}}.
\begin{table*}[!t]
    \centering
   \caption{Anomaly detection and localization results in terms of image/pixel level AUROC on MVTec AD dataset \cite{MVTEC} under the \textbf{unified} setting.}
\label{table:unified}
\fontsize{10}{14}\selectfont{
\resizebox{\textwidth}{!}{\begin{tabular}{cc|ccccccccc|>{\columncolor{orange!15}}cc}
\toprule[1.0pt]
\multicolumn{2}{c|}{\multirow{2}{*}{Category}}             & US        & PSVDD     & PaDiM     & MKD       & DRAEM     & SimpleNet & PatchCore & RD4AD     & UniAD     & \textbf{Ours} & \textbf{Ours} \\
\multicolumn{2}{c|}{} & \cite{ST}           & \cite{PatchSVDD}            & \cite{PaDiM}           & \cite{MKD}           & \cite{draem}          & \cite{liu2023simplenet}           & \cite{PacthCore}           & \cite{RD4AD}           & \cite{uniad}           &($\tau=0.1$) & ($\tau=0.9$)      \\ \midrule
\multicolumn{1}{c|}{\multirow{5}{*}{\rotatebox{90}{Texture}}} & Carpet     & 86.6/88.7 & 97.4/98.5 & 93.8/97.6 & 69.8/95.5 & 98.0/98.6 & 95.9/92.4 & 97.0/98.1 & 97.1/\uline{98.8} & \textbf{99.8}/98.5 &\uline{99.0}/\uline{98.8}  & \uline{99.0}/\textbf{99.0}      \\
\multicolumn{1}{c|}{}                         & Grid       & 69.2/64.5 & 99.1/97.5 & 73.9/71.0 & 83.8/82.3 & \uline{99.3}/\uline{98.7} & 49.8/46.7 & 91.4/98.4 & \textbf{99.7}/\textbf{99.2} & 98.2/96.5 &96.8/98.3 & 98.7/98.3     \\
\multicolumn{1}{c|}{}                         & Leather    & 97.2/95.4 & 95.1/98.1 & \uline{99.9}/84.8 & 93.6/96.7 & 98.7/97.3 & 93.9/96.9 & \textbf{100}/\uline{99.2}  & \textbf{100}/\textbf{99.4}  & \textbf{100}/98.8  &\textbf{100}/99.0 &  \textbf{100}/99.1    \\
\multicolumn{1}{c|}{}                         & Tile       & 93.7/82.7 & 97.3/82.5 & 93.3/80.5 & 89.5/85.3 & \uline{99.8}/\textbf{98.0} & 93.7/93.1 & 96.0/90.3 & 97.5/\uline{95.6} & 99.3/91.8 & \textbf{100}/95.1 & 99.6/94.5     \\
\multicolumn{1}{c|}{}                         & Wood       & 90.6/83.3 & \textbf{99.8}/92.6 & 98.4/89.1 & 93.4/80.5 & \textbf{99.8}/\textbf{96.0} & 95.2/84.8 & 93.8/90.8 & 99.2/\textbf{96.0} & 98.6/93.2 & 98.1/\uline{94.5} & \uline{99.7}/93.8     \\ \midrule
\multicolumn{1}{c|}{\multirow{10}{*}{\rotatebox{90}{Object}}} & Bottle     & 84.0/67.9 & 85.5/86.7 & 97.9/96.1 & 98.7/91.8 & 97.5/87.6 & 97.7/91.2 & \textbf{100}/97.4  & 98.7/97.7 & \uline{99.7}/98.1 & \textbf{100}/\uline{98.4}&\textbf{100}/\textbf{98.5}      \\
\multicolumn{1}{c|}{}                         & Cable      & 60.0/78.3 & 64.4/62.2 & 70.9/81.0 & 78.2/89.3 & 57.8/71.3 & 87.6/88.1 & 95.3/93.6 & 85.0/83.1 & 95.2/97.3 &\textbf{98.4}/\textbf{98.7}  & \uline{97.7}/\uline{98.5}      \\
\multicolumn{1}{c|}{}                         & Capsule    & 57.6/85.5 & 61.3/83.1 & 73.4/96.9 & 68.3/88.3 & 65.3/50.5 & 78.3/89.7 & \textbf{96.8}/98.0 & \uline{95.5}/\uline{98.5} & 86.9/\uline{98.5} & 89.5/\textbf{98.9} & 94.0/\textbf{98.9}     \\
\multicolumn{1}{c|}{}                         & Hazelnut   & 95.8/93.7 & 83.9/97.4 & 85.5/96.3 & 97.1/91.2 & 93.7/96.9 & 99.2/95.7 & 99.3/97.6 & 87.1/\textbf{98.7} & \uline{99.8}/98.1 & \textbf{100}/98.5 & \textbf{100}/\uline{98.6}    \\
\multicolumn{1}{c|}{}                         & Metal Nut  & 62.7/76.6 & 80.9/96.0 & 88.0/84.8 & 64.9/64.2 & 72.8/62.2 & 85.1/90.9 & 99.1/96.3 & \textbf{99.4}/94.1 & \uline{99.2}/94.8 & 98.7/\textbf{97.4} & 98.7/\uline{97.1}     \\
\multicolumn{1}{c|}{}                         & Pill       & 56.1/80.3 & 89.4/96.5 & 68.8/87.7 & 79.7/69.7 & 82.2/94.4 & 78.3/89.7 & 86.4/90.8 & 52.6/96.5 & \uline{93.7}/95.0 & 93.6/\textbf{98.2}& \textbf{94.7}/\uline{97.9}  \\
\multicolumn{1}{c|}{}                         & Screw      & 66.9/90.8 & 80.9/74.3 & 56.9/94.1 & 75.6/92.1 & 92.0/95.5 & 45.5/93.7 & 94.2/98.9 & \textbf{97.3}/\textbf{99.4} & 87.5/98.3 &83.5/98.9 & \uline{95.5}/\uline{99.2}     \\
\multicolumn{1}{c|}{}                         & Toothbrush & 57.8/86.9 & \uline{99.4}/98.0 & 95.3/95.6 & 75.3/88.9 & 90.6/97.7 & 94.7/97.5 & \textbf{100}/98.8  & \uline{99.4}/\textbf{99.0} & 94.2/98.4 &  93.6/98.7 & 98.6/\uline{98.9}    \\
\multicolumn{1}{c|}{}                         & Transistor & 61.0/68.3 & 77.5/78.5 & 86.6/92.3 & 73.4/71.7 & 74.8/64.5 & 82.0/86.0 & 98.9/92.3 & 92.4/86.4 & \uline{99.8}/\uline{97.9} &  98.2/96.0& \textbf{99.9}/\textbf{98.7}    \\
\multicolumn{1}{c|}{}                         & Zipper     & 78.6/84.2 & 77.8/95.1 & 79.7/94.8 & 87.4/86.1 & \uline{98.8}/\uline{98.3} & \textbf{99.1}/97.0 & 97.1/95.7 & 99.6/98.1 & 95.8/96.8 &98.1/\textbf{98.4} & 98.2/\textbf{98.4}    \\ \midrule
\multicolumn{2}{c|}{Mean}                                  & 74.5/81.8 & 76.8/85.6 & 84.2/89.5 & 81.9/84.9 & 88.1/87.2 & 85.1/88.9 & 96.4/95.7 & 93.4/96.0 & 96.5/96.8 & \uline{96.6}/\uline{97.9} & \textbf{98.3}/\textbf{98.0}     \\ \bottomrule[1.0pt]
\end{tabular}}}
\end{table*}
\begin{table}[!h]
\centering
\caption{Anomaly detection and localization results in terms of image/pixel level AUROC on MVTec AD dataset \cite{MVTEC} under the \textbf{few-shot} setting. k represents the number of samples.} 
\label{table:fewshot}
\fontsize{10}{14}\selectfont
\resizebox{\linewidth}{!}{
\begin{tabular}{c|ccccc|>{\columncolor{orange!15}}c}
\toprule[1.0pt]
\multirow{2}{*}{k} & TDG    & DiffNet & PaDiM$^\dag$ & UniAD$^\dag$ & RegAD     & \textbf{Ours} \\
                   & \cite{tdg}        & \cite{differnet} & \cite{PaDiM} &\cite{uniad}        & \cite{RegAD}          & ($\tau=0.1$)     \\ \midrule
2             & 73.2/- & 80.6/- &76.4/89.9 & 81.9/91.3 & \uline{85.7}/\uline{94.6} & \textbf{88.4}/\textbf{95.9}      \\
4             & 74.4/- & 81.3/- &79.6/91.6 & 86.1/93.5  & \uline{88.2}/\uline{95.8} & \textbf{91.9}/\textbf{96.8}      \\
8             & 76.6/- & 83.2/- &82.8/93.3 & 88.8/94.5 & \uline{91.2}/\uline{96.7} & \textbf{93.5}/\textbf{97.8}     \\ \bottomrule[1.0pt]
\end{tabular}
}
\end{table}
\begin{table*}[!h]
\centering
\caption{Comprehensive comparative analysis of the proposed FSR method and existing approaches on MVTec AD \cite{MVTEC} and BTAD \cite{BTAD} datasets. The best result of the existing methods is indicated by a \uwave{wavy underline}.}
\label{table:universal}
\fontsize{10}{14}\selectfont
\resizebox{\linewidth}{!}{\begin{tabular}{c|c|cccccc|>{\columncolor{orange!15}}cccc}
\toprule[1.0pt]
  \multicolumn{2}{c|}{\multirow{2}{*}{Setting}} & PaDiM & DRAEM & FastFlow & PatchCore & RD4AD & UniAD & \textbf{Ours} & \textbf{Ours} & \textbf{Ours} & \textbf{Ours} \\
                \multicolumn{2}{c|}{} & \cite{PaDiM}      & \cite{draem}       & \cite{fastflow}         & \cite{PacthCore}          & \cite{RD4AD}       & \cite{uniad}      & ($\tau=0.1$)              & ($\tau=0.3$)              &  ($\tau=0.9$)             &  (Optimal)             \\ \midrule 
 \multicolumn{1}{c|}{\multirow{4}{*}{\rotatebox{90}{MVTec AD}}} & Separate          & 95.5/97.5  & 98.0/97.3  & 97.5$^\dag$/\uwave{98.4$^\dag$}      & \uwave{99.1}/98.1      & 98.5/97.8  & 96.6/96.6  & \uline{99.1}/\textbf{98.4}  & \textbf{99.2}/\textbf{98.4}               & 98.9/\uline{98.3}              & \textbf{99.2}/\textbf{98.4}               \\
\multicolumn{1}{c|}{} & Unified           & 84.2/89.5  & 88.1/87.2  & 91.1$^\dag$/96.2$^\dag$          & 96.4/95.7      & 93.4/96.0  & \uwave{96.5}/\uwave{96.8}  & 96.6/\uline{97.9}              & \uline{97.1}/\textbf{98.0}              & \textbf{98.3}/\textbf{98.0}               & \textbf{98.3}/\textbf{98.0}              \\
\multicolumn{1}{c|}{} & Few-shot-8        & 82.8$^{\dag}$/93.3$^{\dag}$      &  81.2$^{\dag}$/91.5$^{\dag}$     & \uwave{92.4}$^{\dag}$/\uwave{97.0}$^{\dag}$         &  92.1$^{\dag}$/96.5$^{\dag}$         & 90.5$^{\dag}$/95.9$^{\dag}$     & 88.8$^{\dag}$/94.5$^{\dag}$      & \textbf{93.5}/\textbf{97.8}               & \uline{92.5}/\uline{97.4}               & 90.4/96.6               & \textbf{93.5}/\textbf{97.8}              \\ 
\cmidrule(lr){2-12}
\multicolumn{1}{c|}{} & Mean              & 87.5/93.4      & 89.1/92.0      & 93.7/\uwave{97.2}         & \uwave{95.9}/96.8          & 94.1/96.6      & 94.0/96.0     & \uline{96.4}/\uline{98.0}              & 96.3/97.9              & 95.9/97.6               & \textbf{97.0}/\textbf{98.1}              \\ \midrule
 \multicolumn{1}{c|}{\multirow{4}{*}{\rotatebox{90}{BTAD}}} & Separate          & 93.9$^\dag$/97.2$^\dag$  & 89.0/87.1  & 90.1/96.3      & 94.3$^\dag$/96.9$^\dag$      & \uwave{94.4}$^\dag$/\uwave{97.5}$^\dag$  & 93.8$^\dag$/97.0$^\dag$  & \uline{95.6}/\textbf{97.6}  & \textbf{95.9}/97.4               & 95.5/\uline{97.5}              & \textbf{95.9}/\textbf{97.6}               \\
\multicolumn{1}{c|}{} & Unified           & 45.4$^\dag$/82.9$^\dag$  & 88.1$^\dag$/86.8$^\dag$  & 81.4$^\dag$/88.7$^\dag$          & \uwave{94.6}$^\dag$/97.1$^\dag$      & 93.9$^\dag$/97.1$^\dag$  & 92.4$^\dag$/\uwave{97.2}$^\dag$  & 95.2/\uline{97.2}              & \uline{95.4}/\uline{97.2}              & \textbf{95.5}/\textbf{97.3}               & \textbf{95.5}/\textbf{97.3}              \\
\multicolumn{1}{c|}{} & Few-shot-8        & \uwave{92.4}$^{\dag}$/\uwave{96.9}$^{\dag}$      &  83.8$^{\dag}$/79.0$^{\dag}$     & 89.1$^{\dag}$/95.3$^{\dag}$         &  91.8$^{\dag}$/94.9$^{\dag}$         & 92.0$^{\dag}$/96.6$^{\dag}$     & 92.0$^{\dag}$/96.7$^{\dag}$      & \uline{93.0}/\textbf{97.1}               & \textbf{93.2}/\uline{97.0}               & 91.4/96.6               & \textbf{93.2}/\textbf{97.1}              \\ 
\cmidrule(lr){2-12}
\multicolumn{1}{c|}{} & Mean              & 77.2/92.3      & 87.0/84.3      & 86.9/93.4         & \uwave{93.6}/96.3          & 93.4/\uwave{97.1}      & 92.7/97.0     & 94.6/\textbf{97.3}              & \uline{94.8}/\uline{97.2}              & 94.1/97.1               & \textbf{94.9}/\textbf{97.3}              \\ \midrule
\multicolumn{1}{c|}{{\multirow{3}{*}{\rotatebox{90}{Efficency}}}} & Params(M)               & 1310.77       & 97.42      & 69.84         & \uline{68.88}         &150.64       & \textbf{26.48}       & 125.64                & 125.64             & 125.64               & 125.64              \\ 
\multicolumn{1}{c|}{} & FLOPs(G)               & 30.37       & 198.66      & 36.24         & \uline{24.13}         &38.94       & \textbf{8.50}       & 37.85                & 37.85             & 37.85             & 37.85             \\ 
\multicolumn{1}{c|}{} & Inf. time(ms)               & 155.04       & \textbf{16.13}      & 37.74         & 89.85         &30.59      & 57.54       & \uline{24.44}                & \uline{24.44}              & \uline{24.44}               & \uline{24.44}              \\ \bottomrule[1.0pt]
\end{tabular}}
\end{table*}
\begin{figure*}[!h]
    \centering
    \includegraphics{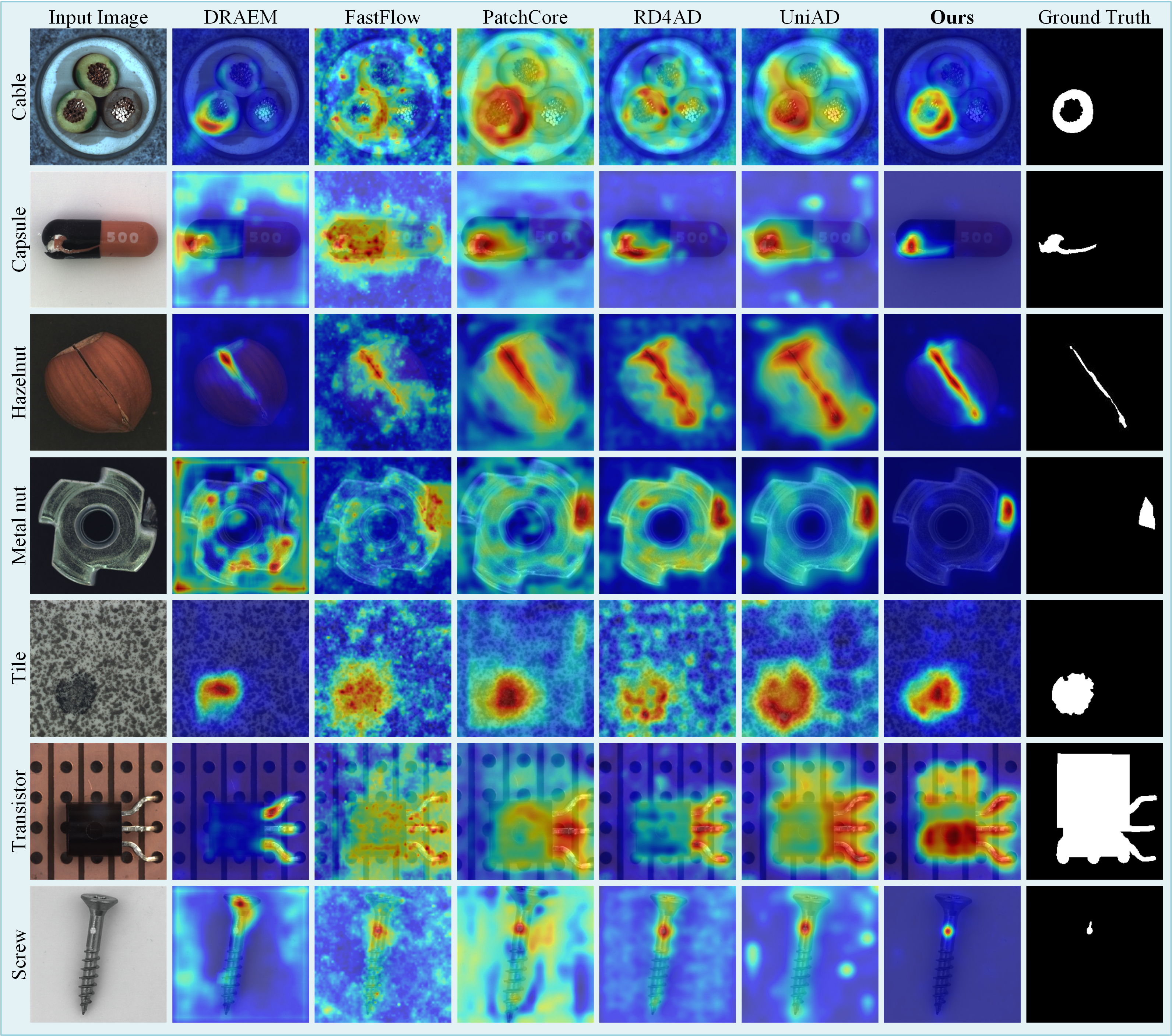}
    \caption{The qualitative segmentation results of the proposed model and the comparative methods under the unified setting on MVTec AD \cite{MVTEC} benchmark. In contrast to alternative approaches, our method exhibits superior  segmentation accuracy across various samples and defect types.}
    \label{fig:mvtec-sample}
    \vspace{-1em}
\end{figure*}
\begin{figure*}
    \centering
    \includegraphics{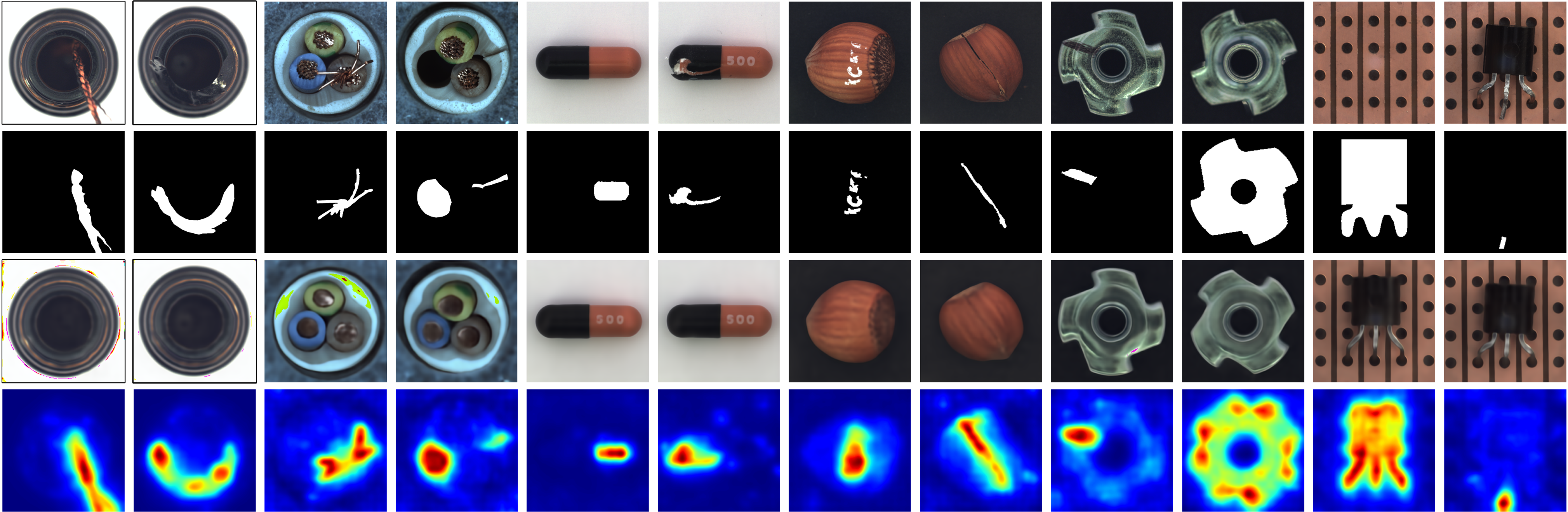}
    \caption{{The reconstruction and anomaly segmentation results of our model. From top to bottom: input anomalous image, ground truth, reconstructed image, and predicted anomaly map. Notably, we utilize an additional decoder to visualize the reconstructed features as images. This decoder is solely designed for visualization purposes.}}
    \label{fig:add_example}
\end{figure*}
\begin{table}[]
\centering
\caption{{Comprehensive comparison with the latest SOTA methods on the MVTec-AD~\cite{MVTEC} dataset.}}
\fontsize{10}{14}\selectfont{
\resizebox{\linewidth}{!}{
\begin{tabular}{c|ccc|>{\columncolor{orange!15}}cc}
\toprule[1.0pt]
Method~$\rightarrow$ & FastRecon~\cite{fang2023fastrecon} &  FOD~\cite{yao2023focus}       & DiAD~\cite{he2024diffusion}      & \textbf{Ours}      & \textbf{Ours}      \\
                       Setting~$\downarrow$  & ICCV'23                  & ICCV'23          & AAAI'24          & ($\tau=0.1$)         & (Optimal)         \\ \midrule
Separate                 & 95.7$^\dag$/97.8$^\dag$  & \textbf{99.2}/\uline{98.3} & 90.8$^\dag$/95.7$^\dag$ & \uline{99.1}/\textbf{98.4} & \textbf{99.2}/\textbf{98.4} \\
Unified                  & 72.0$^\dag$/87.5$^\dag$        & 92.6$^\dag$/\textbf{98.0}$^\dag$ & \uline{97.2}/96.8 & 96.6/\uline{97.9} & \textbf{98.3}/\textbf{98.0} \\
Few-shot-8               & \textbf{95.2}/\uline{97.3}        & 83.4$^\dag$/94.5$^\dag$ & 85.0$^\dag$/92.2$^\dag$ & \uline{93.5}/\textbf{97.8} & \uline{93.5}/\textbf{97.8} \\ \midrule
Mean                     & 87.6/94.2       & 95.1/97.0 & 91.0/94.9 & \uline{96.4}/\uline{98.0} & \textbf{97.0}/\textbf{98.1} \\ \bottomrule[1.0pt]
\end{tabular}}}
\label{table:comparedwithsota}
\end{table}
\begin{figure}[!h]
    \centering
    \includegraphics[width=85mm]{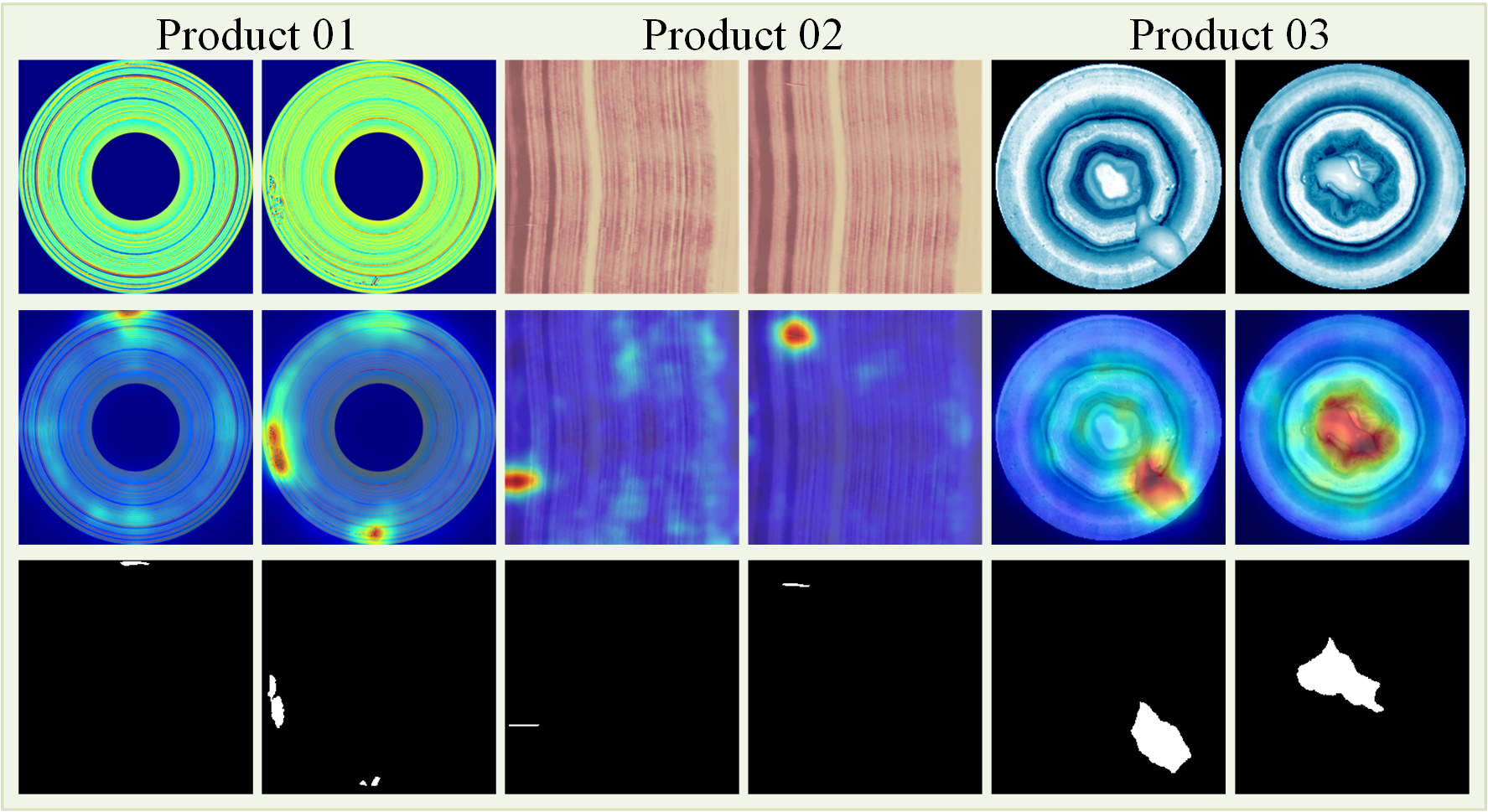}
    \caption{The qualitative segmentation results of proposed method on BTAD \cite{BTAD} dataset. Top row: The anomalous image. Mid row:
The anomaly scoring map. Bottom row: The ground truth.}
    \label{fig:BTAD}
\end{figure}
\subsection{Anomaly detection on MVTec AD}
\subsubsection{Comparison with the baseline model}
The quantitative experimental results on MVTec AD under the separate, unified, and few-shot settings are presented in Tables \ref{table:separate}, \ref{table:unified}, and \ref{table:fewshot}, respectively. Across all three settings, our approach consistently achieves superior detection and localization results. In terms of the image/pixel AUROC indicators, our method outperforms PatchCore by 0.1/0.3\% in the separate setting, surpasses UniAD by 1.8/1.2\% in the unified setting, and exceeds RegAD by 2.3/1.1\% in the few-shot ($k=8$) setting. \\
\indent A more comprehensive comparison of results is presented in Table \ref{table:universal}. Despite the outstanding detection performance exhibited by previous methods in specific settings, their detection performance sharply declines when transferred to different settings. For instance, PatchCore, known for its excellent performance in the separate setting, undergoes a decrease of 2.7/2.4\% in image/pixel AUROC, respectively, when transitioned to the unified setting. In contrast, our method (optimal) demonstrates a remarkably robust performance during the transition from the separate setting to the unified setting, experiencing only a marginal decrease of 0.9/0.4\% in image/pixel AUROC, respectively. Similarly, UniAD, known for its excellent performance in the unified setting, experiences a decrease of 7.7/2.3\% in image/pixel AUROC, respectively, when transferred to the few-shot setting. In contrast, our method (optimal) exhibits a more robust performance when transitioning from the unified setting to the few-shot setting, with only the decrease of 4.8/0.2\% in image/pixel AUROC. The RegAD method, which exhibits excellent performance in the few-shot setting, is unable to be seamlessly transferred to alternative settings due to its intricate configuration.\\
\indent Our method (optimal) achieves the best average results, with an image AUROC of 97.0\% and a pixel AUROC of 98.1\%, surpassing the best results of the previous methods by 1.1\% and 0.9\%, respectively. The optimal results of our method are obtained using different shuffling rates under different settings. Such comparisons may seem unfair. Therefore, we also report the performance of our method at a fixed shuffling rate. Our approach ($\tau=0.1$) surpasses the best results of previous methods by 0.5\% and 0.8\%, respectively.\\
\indent The qualitative comparison experimental results are shown in Fig. \ref{fig:mvtec-sample}. Our proposed method accurately localizes defect regions and suppresses background noise. In contrast, the anomaly localization maps obtained by UniAD and PatchCore are notably coarse and exhibit significant noise, attributable to their utilization of low-resolution feature maps. Moreover, our method demonstrates strong robustness, delivering outstanding localization results across various categories. In contrast, DRAEM method performs well on the Tile category but poorly on the Transistor category, lacking robustness. Finally, our method demonstrates significant advantages in detecting logical defects, such as misplaced defects in the Transistor category and cable swap defects in the Cable category. This is attributed to our proposed FSR strategy, which enables the model to learn global semantic information.\\
\indent {Fig.~\ref{fig:add_example} presents additional reconstruction and anomaly segmentation results of our method. As shown, our approach effectively addresses the identical shortcut problem, avoiding the direct copying of anomalous patterns. It successfully restores anomalous patterns to normal ones, enabling precise segmentation of the anomalous regions.}\\
\indent In addition, achieving a balance between detection speed and accuracy is crucial in real industrial scenarios. Therefore, we also report the Average Inference Time, Parameters, and Floating Point Operations (FLOPs) of our method and existing methods in Table \ref{table:universal}. Our approach achieves the second-highest detection speed, with an average inference time of 24.44ms, only slightly trailing behind DRAEM with an average inference time of 16.13ms. However, our method significantly outperforms DRAEM in terms of detection accuracy. The previous SOTA PatchCore achieves an average inference time of only 89.85ms, whereas our approach is almost \textbf{4$\times$} faster than PatchCore. The parameters and FLOPs of our method are 125.64M and 37.85G, respectively, which are lower than RD4AD's parameters and DRAEM's FLOPs. However, our method's detection performance far exceeds that of RD4AD and DRAEM. This indicates that the performance improvement of our method does not come from a larger model capacity.
\subsubsection{{Comparison with Recent SOTA models}}
{To further validate the effectiveness of our method, we compare it with several recently proposed state-of-the-art (SOTA) methods, including FastRecon~\cite{fang2023fastrecon}, which achieves SOTA performance in the few-shot setting, FOD~\cite{yao2023focus}, which performs SOTA in the separate setting, and DiAD~\cite{he2024diffusion}, which excels in the unified setting. However, these methods are designed for specific settings, and to conduct a fair comparison, we use their officially released code\footnote{\href{https://github.com/FzJun26th/FastRecon}{\textcolor{codered}{https://github.com/FzJun26th/FastRecon}}}\footnote{\href{https://github.com/xcyao00/FOD}{\textcolor{codered}{https://github.com/xcyao00/FOD}}}\footnote{\href{https://github.com/lewandofskee/DiAD}{\textcolor{codered}{https://github.com/lewandofskee/DiAD}}} to reproduce their performance in other settings.\\
\indent Table~\ref{table:comparedwithsota} presents the results of this experiment. FastRecon excels in the few-shot setting, but its performance drops significantly when transferred to the unified setting. This is because FastRecon models the data distribution as a single Gaussian, a hypothesis that does not hold in the unified setting. Similarly, FOD and DiAD achieve strong performance in their respective separate and unified settings but suffer a notable decline in the few-shot setting. This is due to the large number of meticulously designed modules in both methods, which are difficult to fully train and fit with a limited number of samples. In contrast, our method (Optimal) achieves the best or second-best performance across all settings. More importantly, in our default setting ($\tau=0.1$), our method significantly outperforms these approaches in average performance across the three settings, thus fully demonstrating the superiority of our approach.}

\begin{table}[]
\caption{{Performance of our method under different settings across multiple random seeds on the MVTec-AD~\cite{MVTEC} dataset.}}
\fontsize{10}{12}\selectfont{
\resizebox{0.85\linewidth}{!}{
\begin{tabular}{c|ccc}
\toprule[1.0pt]
Setting~$\rightarrow$ & \multirow{2}{*}{Unified} & \multirow{2}{*}{Seperate} & \multirow{2}{*}{Few-Shot-4} \\
Seed~$\downarrow$    &                          &                           &                             \\ \midrule
1       & 96.6/97.9                & 99.1/98.4                 & 91.9/96.8                   \\
42      & 96.6/97.9                & 99.1/98.4                 & 91.8/96.8                   \\
123     & 96.5/97.8                & 99.1/98.5                 & 91.7/96.9                   \\
2025    & 96.7/97.9                & 99.1/98.4                 & 91.8/96.8                   \\ \midrule
Mean    & 96.60/97.88              & 99.10/98.43               & 91.80/96.83                 \\
Std     & 0.08/0.05              & 0.00/0.05               & 0.08/0.05                 \\ \bottomrule[1.0pt]
\end{tabular}}}
\label{table:seed}
\end{table}
\subsubsection{{Stability Evaluation with Multiple Random Seeds}}
{The random shuffling operation introduces inherent stochasticity during training, which may affect model performance. To systematically evaluate the stability of the FSR method, we conducted experiments using multiple random seeds across three different settings. As shown in Table~\ref{table:seed}, the model exhibits highly consistent performance across four seeds, with standard deviations below 0.1\% for all metrics. These results demonstrate that FSR is robust to the stochasticity introduced during training, confirming its reliability and providing strong evidence for its practical applicability.}
\subsection{Anomaly detection on BTAD}
The comprehensive quantitative comparison experiment results on the BTAD dataset are showcased in Table \ref{table:universal}. Our approach ($\tau=0.1$) achieves optimal detection results across the few-shot, separate, and unified settings. It demonstrates improvements in image/pixel AUROC indicators by 0.6/0.2\%, 1.2/0.1\%, and 0.6/0.1\%, respectively, compared to the SOTA results of previous methods. Taking into account all three settings, our method achieves superior detection performance, with an image AUROC of 94.6\% and a pixel AUROC of 97.3\%. Compared to the second-best results, our method shows improvements of 1.0\% and 0.2\%, respectively. This highlights the adaptability of our proposed method to various industrial settings, demonstrating significant potential for industrial applications.\\
\indent The qualitative experimental results are depicted in Fig. \ref{fig:BTAD}. Our method is capable of precisely localizing anomalies of various sizes even in complex textured backgrounds.
\begin{figure}
    \centering
    \includegraphics[width=85mm]{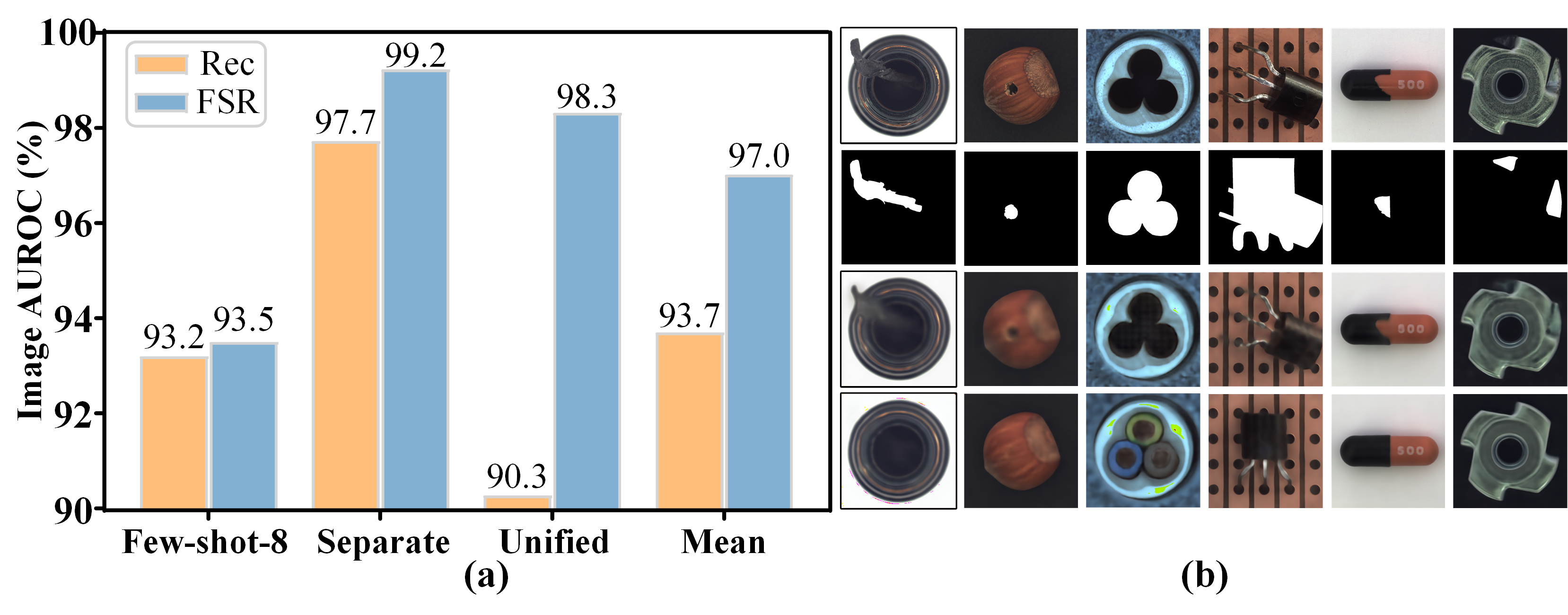}
    \caption{The impact of FSR strategy. (a) Quantitative ablation results. (b) Qualitative ablation results. First row: The anomalous image. Second row: The ground truth. Third row: The reconstruction result \textbf{without} FSR strategy. Last row: The reconstruction result \textbf{with} FSR strategy.}
    \label{fig:ablation_FSR}
\end{figure}
\begin{figure}[!t]
    \centering
    \includegraphics[width=85mm]{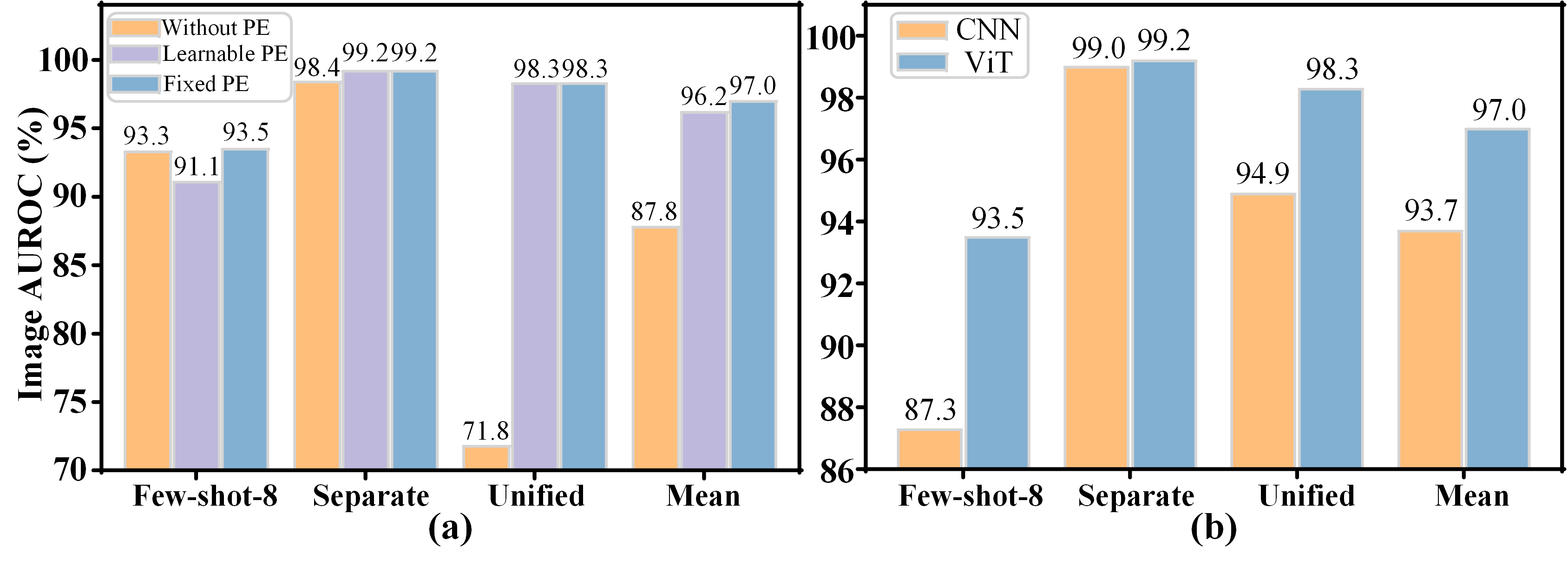}
    \caption{(a) The ablation results regarding positional embedding (PE). (b) The ablation results regarding Vision Transformer (ViT).}
    \label{fig:ablation_posandvit}
\end{figure}

\begin{figure*}[!ht]
    \centering
    \includegraphics[width=180mm]{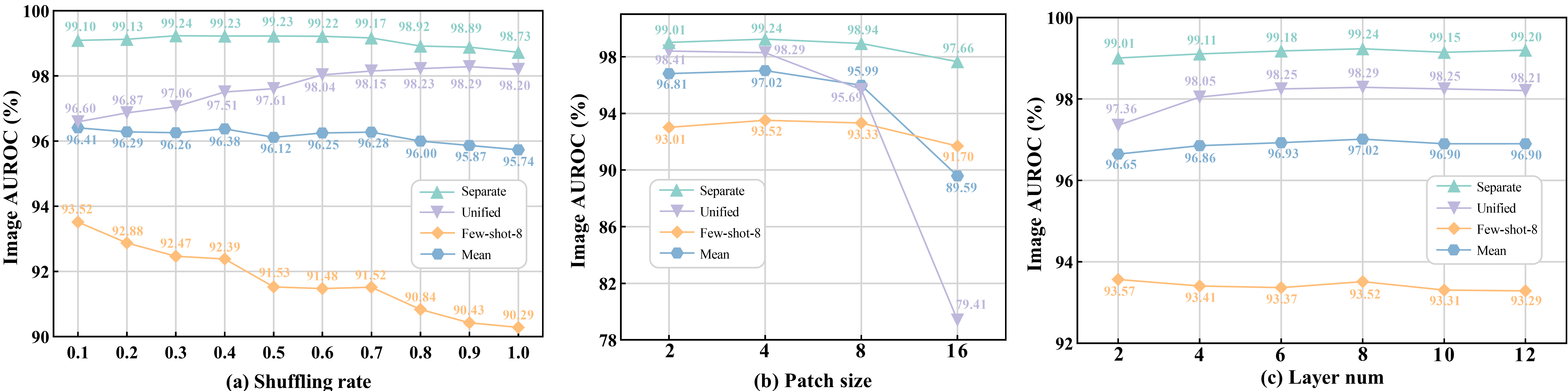}
    \caption{Ablation studies with image AUROC metric on MVTec AD \cite{MVTEC}. (a) Anomaly detection performance with different shuffling rate $\tau$. (b) Anomaly detection performance with different patch size $P$. (c) Anomaly detection performance with different layer number $N$.}
    \label{fig:ablation}
\end{figure*}
\subsection{Ablation study}
\subsubsection{{Influence of the FSR strategy}}
{Fig. \ref{fig:ablation_FSR} illustrates the  significance of FSR strategy. As depicted in Fig. \ref{fig:ablation_FSR}(a), compared to the model optimized with Rec task, the model optimized with FSR strategy exhibits enhancements in image AUROC metric of 0.3\%, 1.5\%, and 8.0\% in the few-shot, separate, and unified settings, respectively. Considering all three settings collectively, the FSR strategy results in a 3.3\% improvement in Image AUROC. Further, we observe that as the transition progresses from a few-shot setting to a unified setting, the performance improvement brought by the FSR strategy becomes more significant. This is attributed to the escalating complexity of the normal data distribution from the few-shot setting to the unified setting, thereby exacerbating the issue of identical shortcut in the Rec proxy task. Our FSR strategy effectively mitigates identical shortcut issue across various settings. {The qualitative ablation results are depicted in Fig. \ref{fig:ablation_FSR}(b). The model without the FSR strategy still reconstructs anomalies, whereas the model with the FSR strategy can rectify anomalies into normal patterns, providing a more intuitive validation of the effectiveness of the FSR strategy.}}
\subsubsection{Influence of the positional embedding}
Fig. \ref{fig:ablation_posandvit}(a) illustrates the impact of positional embedding on the detection performance of our model. Compared to the default setup (fixed sinusoidal positional embedding), the model without positional embedding experiences a decline in detection performance of 0.2\%, 0.8\%, and 26.5\% in the few-shot, separate, and unified settings, respectively. The sharp decline in detection performance under the unified setting is attributed to the training dataset comprising images of various products, necessitating positional embedding to aid the model in better establishing the distribution of normal data. Furthermore, the model using learnable positional embedding do not exhibit performance degradation in the separate and unified settings. However, there is a significant decline in detection performance of 2.4\% in the few-shot setting. This is attributed to the limited number of samples in the training dataset under the few-shot setting, which is insufficient for learning representative positional embedding. Hence, in our study, we recommend employing fixed sinusoidal positional embedding.
\subsubsection{Influence of the Vision Transformer}
The selection of the restoration network is pivotal for FSR task, as it necessitates the ability for global modeling. In our approach, we utilize Vision Transformer (ViT) as the restoration network. To explore its impact, we conducted ablation experiments by replacing ViT with a CNN of equivalent parameter size. As depicted in Fig. \ref{fig:ablation_posandvit}(b), after replacing ViT with CNN, the model's detection performance decreases in the few-shot, separate, and unified settings by 6.2\%, 0.2\%, and 3.4\%, respectively. This is attributed to the inherent structure of CNN, which restricts its consideration to the semantic information within fixed convolutional kernels, thus lacking the capability for global modeling. In contrast, ViT's multi-head attention mechanism considers the interrelations between each feature block, providing the ability for global modeling, thus making it highly suitable for FSR task.

\subsubsection{{Analysis of Model Hyperparameters}}
\label{effectshuffling}
{
Fig. \ref{fig:ablation}(a) illustrates the relationship between shuffling rate $\tau$ and detection performance under different settings. The shuffling rate controls the difficulty of the agent task. As shown in Fig. \ref{fig:ablation}(a), the optimal shuffling rate varies in different settings, with 0.1 in the few-shot setting, 0.3 in the separate setting, and 0.9 in the unified setting. This is because from the few-shot setting, to the separate setting, and to the unified setting, the more complex the normal data distribution becomes, the more serious the identical shortcut problem becomes. Therefore, in the few-shot setting where the identical shortcut problem is less severe, the relatively easy FSR task produced using a shuffling rate of 0.1 allows the model to exhibit the best detection performance. In contrast, in the unified setting where the identical shortcut problem is very severe, the difficult FSR task generated using a shuffling rate of 0.9 allows the model to exhibit optimal performance. Overall, in settings where the data distribution is more complex, it is necessary to use a larger shuffling
rate to produce more difficult agent tasks, thus avoiding the model from falling into the identical shortcut problem. Additionally, we also report the relationship between this shuffling rate and the average detection performance in the three settings. The optimal shuffling rate is 0.1 in terms of average detection performance.\\
\indent Fig. \ref{fig:ablation}(b) illustrates the relationship between patch size $P$ and detection performance under different settings. The patch size decides the granularity for divided feature blocks, which affects the difficulty of the FSR task and consequently influences the model's inspection performance. As illustrated in Fig. \ref{fig:ablation}(b), the optimal patch size is 4.\\
\indent Fig. \ref{fig:ablation}(c) illustrates the relationship between layer number $N$ and detection performance under different settings. The layer number determines the network's restorative capacity, thereby influencing the model's detection performance. As depicted in Fig. \ref{fig:ablation}(c), the optimal layer number is 8.}

\begin{figure}[!t]
    \centering
    \includegraphics[width=88mm]{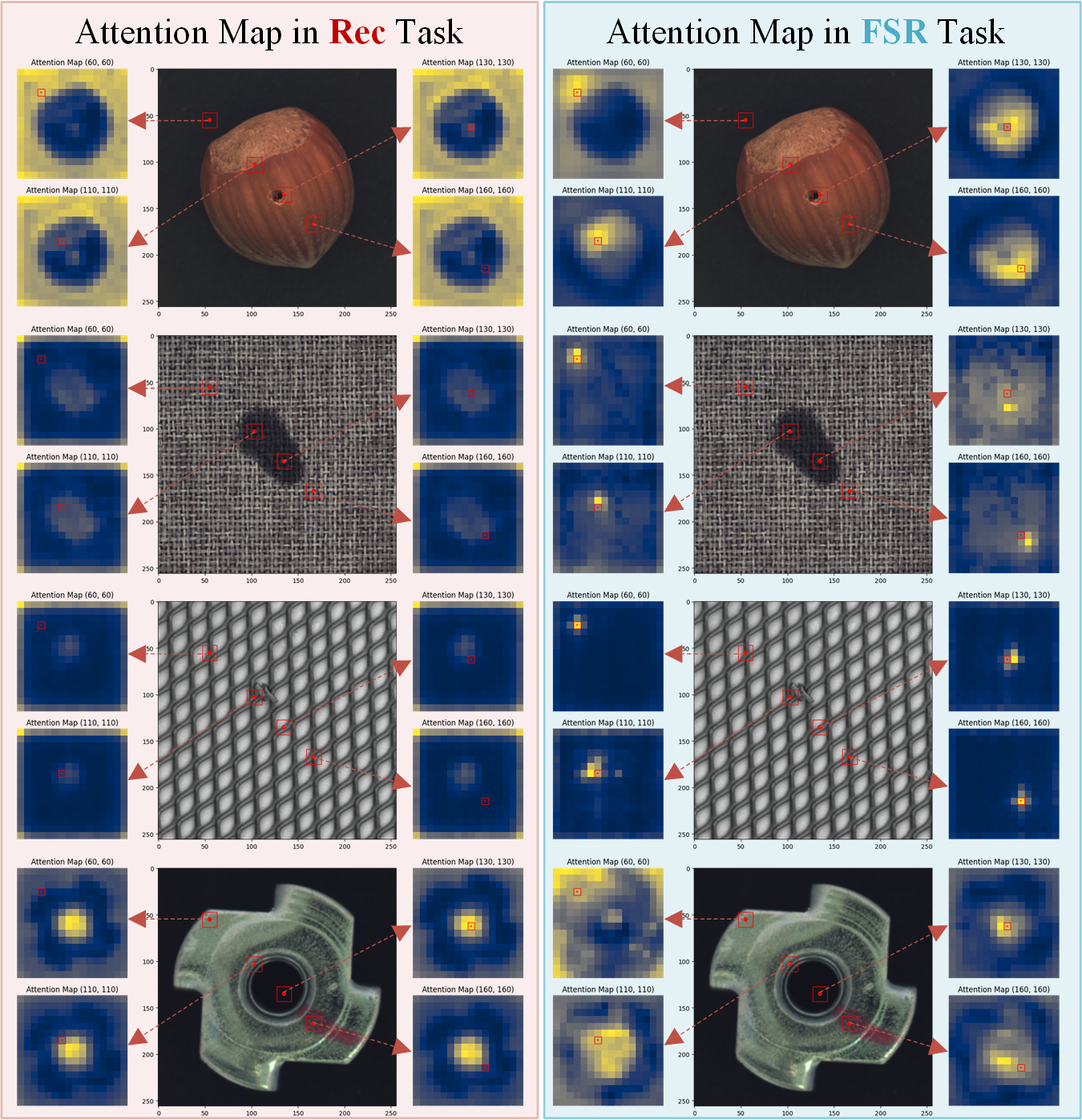}
    \caption{Visualization of attention maps in reconstruction (Rec) task and feature shuffling and restoration (FSR) task. Each {red} box in the figure represents an image block, and we visualize its discrete similarity distribution with all the other image blocks.}
    \label{fig:attenvis}
    \vspace{-0.5em}
\end{figure}
\subsection{Attention map visualization}
{Fig. \ref{fig:attenvis} visualizes the attention maps in Rec task and FSR task. As depicted in the left side of Fig. \ref{fig:attenvis}, we observe that the model trained for Rec task exhibits identical attention maps regardless of the input feature block. This implies that it fundamentally disregards the semantic information of the input feature block, so its $MSA$ module  consistently outputs a constant value. This observation supports the intuitive explanation provided in Section \ref{explanation-ns}. In contrast, as shown in the right side of Fig. \ref{fig:attenvis}, the model trained for FSR task generates distinct attention maps for different feature blocks based on their semantic contents, enabling the output of $MSA$ module to contain more semantically meaningful information. Notably, for normal feature blocks, the model trained for FSR task places greater emphasis on the block itself, while for abnormal feature blocks, it pays more attention to the surrounding similar normal features. This allows the model to leverage global contextual information to restore abnormal features.}
\begin{figure}[!t]
    \centering
    \includegraphics[width=85mm]{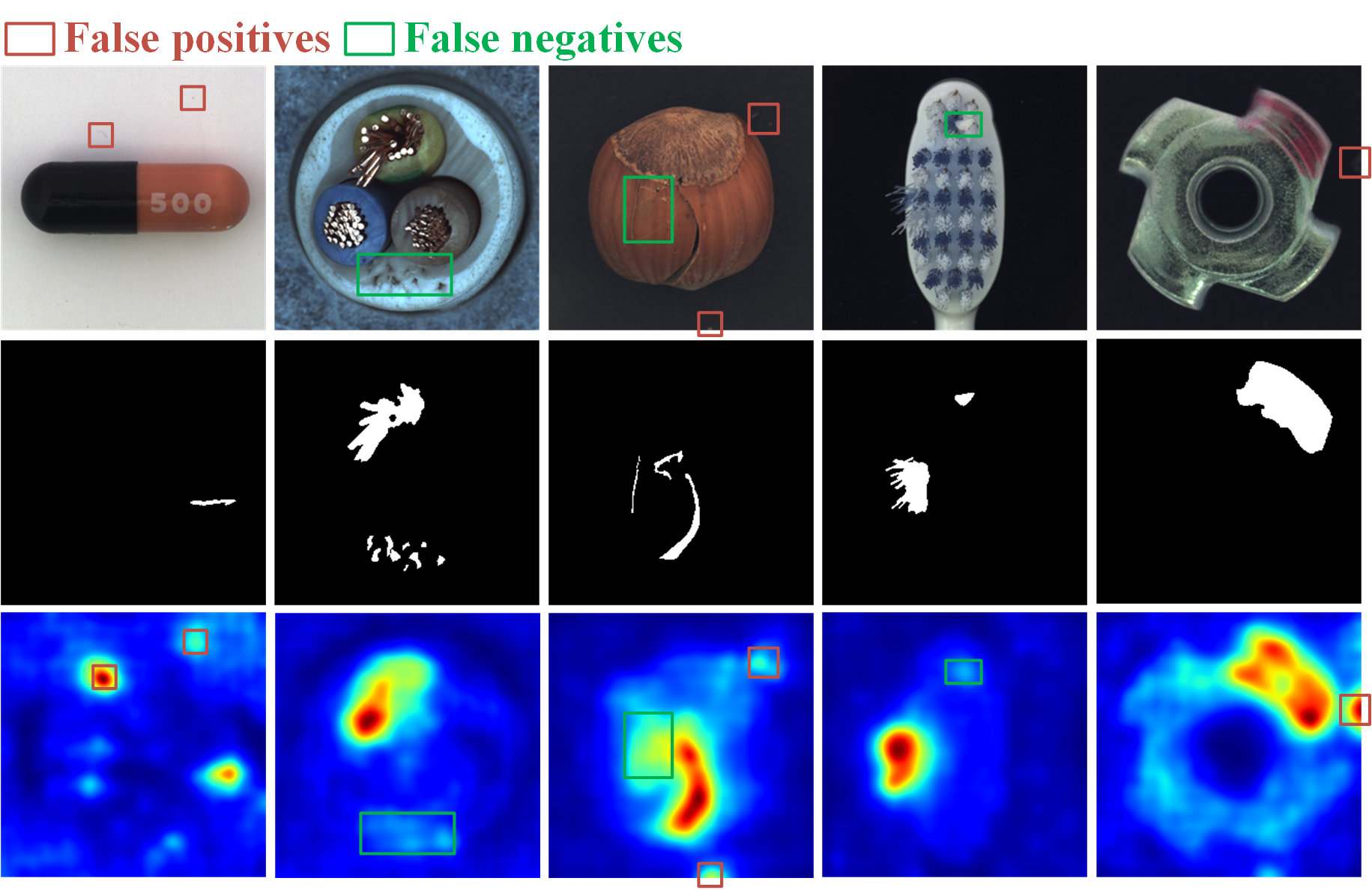}
    \caption{Failure cases of our method. From top to bottom: input anomalous image, ground truth, and predicted anomaly map.}
    \label{fig:failuercase}
\end{figure}
\subsection{Analysis of failure cases}
As shown in Fig.~\ref{fig:failuercase}, although our method effectively addresses the issue of identical shortcut, some failure cases still occur, which can be primarily categorized into two types. The first type involves false positives caused by background noise, while the second type consists of false negatives arising from subtle anomalies or anomalies that resemble the normal background. The underlying reason for these detection failures lies in the limited discriminative power of the pre-trained feature extractor in distinguishing between normal and anomalous patterns. The feature extractor used in our method was pre-trained on the ImageNet~\cite{ImageNet} dataset, whose image distribution differs significantly from that of industrial images. Therefore, future research will focus on enhancing the discriminative capability of the feature extractor for specific industrial images, aiming to further improve detection performance.

\section{Conclusion}
In this paper, we introduce a straightforward and effective unsupervised method, termed FSR, designed for universal anomaly detection. Focusing on the fundamental perspective of agent tasks, FSR addresses identical shortcut issues across diverse settings. The pivotal hyperparameter, shuffling rate, serves to regulate the difficulty of FSR tasks, thereby optimizing model performance for various application scenarios. Under few-shot, separate, and unified settings, our approach consistently achieves state-of-the-art performance on the MVTec AD and BTAD datasets. Furthermore, it attains a favorable balance between detection accuracy and speed, rendering it suitable for real-world industrial applications. \\
\indent While our method demonstrates excellent detection performance in various settings, optimal model performance requires manual tuning of the hyperparameter, shuffling rate. Therefore, our future research is dedicated to exploring adaptive adjustments of the shuffling rate based on different settings, aiming to automate this parameter tuning process for enhanced model adaptability and performance optimization.

\section*{CRediT authorship contribution statement}
\textbf{Wei Luo}: Conceptualization, Methodology, Formal analysis, Investigation, Writing–original draft, Writing–review \& editing, Visualization. 

\textbf{Haiming Yao}: Methodology, Writing–review \& editing, Validation, Supervision. 

\textbf{Zhenfeng Qiang}: Visualization, Methodology, Validation.

\textbf{Xiaotian Zhang}: Methodology, Validation, 

\textbf{Weihang Zhang}: Funding acquisition, Supervision, Writing-review \& editing.

\section*{Declaration of competing interest}
The authors declare that they have no known competing financial interests or personal relationships that could have appeared to
influence the work reported in this paper.
\section*{Data availability}
The data is available online.
\section*{Acknowledgments}
This study was financially supported by the National Natural Science Foundation of China (NSFC) under Grant 62506036 and the Beijing Institute of Technology Research Fund Program for Young Scholars under Grant 6120220236.
\bibliographystyle{elsarticle-num}

\bibliography{cas-refs}

@article{PGANet,
  title={PGA-Net: Pyramid Feature Fusion and Global Context Attention Network for Automated Surface Defect Detection},
  author={Hongwen Dong and Kechen Song and Yu He and Jing Xu and Yunhui Yan and Qinggang Meng},
  journal={IEEE Transactions on Industrial Informatics},
  year={2020}
}

@article{PatchSVDD,
  title={Patch SVDD: Patch-level SVDD for Anomaly Detection and Segmentation},
  author={Jihun Yi and Sungroh Yoon},
  journal={asian conference on computer vision},
  year={2020}
}

@article{ST,
  title={Uninformed Students: Student-Teacher Anomaly Detection with Discriminative Latent Embeddings.},
  author={Paul Bergmann and Michael Fauser and David Sattlegger and Carsten Steger},
  journal={computer vision and pattern recognition},
  year={2019}
}

@article{AE,
  title={Reducing the Dimensionality of Data with Neural Networks},
  author={Geoffrey E. Hinton and Ruslan Salakhutdinov},
  journal={Science},
  year={2006}
}

@article{MemAE,
  title={Memorizing Normality to Detect Anomaly: Memory-Augmented Deep Autoencoder for Unsupervised Anomaly Detection},
  author={Dong Gong and Lingqiao Liu and Vuong Le and Budhaditya Saha and Moussa Reda Mansour and Svetha Venkatesh and Anton van den Hengel},
  journal={international conference on computer vision},
  year={2019}
}

@article{TrustMAE,
  title={TrustMAE: A Noise-Resilient Defect Classification Framework using Memory-Augmented Auto-Encoders with Trust Regions},
  author={Daniel Stanley Tan and Yi-Chun Chen and Trista Pei-Chun Chen and Wei-Chao Chen},
  journal={workshop on applications of computer vision},
  year={2021}
}

@article{RIAD,
  title={Reconstruction by inpainting for visual anomaly detection},
  author={Zavrtanik, Vitjan and Kristan, Matej and Sko{\v{c}}aj, Danijel},
  journal={Pattern Recognition},
  volume={112},
  pages={107706},
  year={2021},
  publisher={Elsevier}
}

@article{Ganomaly,
  title={GANomaly: Semi-Supervised Anomaly Detection via Adversarial Training},
  author={Samet Akcay and Amir Atapour-Abarghouei and Toby P. Breckon},
  journal={asian conference on computer vision},
  year={2018}
}

@article{FMR-Net,
  title={A Feature Memory Rearrangement Network for Visual Inspection of Textured Surface Defects Toward Edge Intelligent Manufacturing},
  author={Yao, Haiming and Yu, Wenyong and Wang, Xue},
  journal={IEEE Transactions on Automation Science and Engineering},
  year={2022},
  publisher={IEEE}
}

@article{MVTEC,
  title={MVTec AD — A Comprehensive Real-World Dataset for Unsupervised Anomaly Detection},
  author={Paul Bergmann and Michael Fauser and David Sattlegger and Carsten Steger},
  journal={computer vision and pattern recognition},
  year={2019}
}

@inproceedings{draem,
  title={Draem-a discriminatively trained reconstruction embedding for surface anomaly detection},
  author={Zavrtanik, Vitjan and Kristan, Matej and Sko{\v{c}}aj, Danijel},
  booktitle={Proceedings of the IEEE/CVF International Conference on Computer Vision},
  pages={8330--8339},
  year={2021}
}

@article{ImageNet,
  title={ImageNet Large Scale Visual Recognition Challenge},
  author={Olga Russakovsky and Jia Deng and Hao Su and Jonathan Krause and Sanjeev Satheesh and Sean Ma and Zhiheng Huang and Andrej Karpathy and Aditya Khosla and Michael S. Bernstein and Alexander C. Berg and Li Fei-Fei},
  journal={International Journal of Computer Vision},
  year={2014}
}

@article{AE-SSIM,
  title={Improving unsupervised defect segmentation by applying structural similarity to autoencoders},
  author={Bergmann, Paul and L{\"o}we, Sindy and Fauser, Michael and Sattlegger, David and Steger, Carsten},
  journal={arXiv preprint arXiv:1807.02011},
  year={2018}
}

@article{DFR,
  title={Unsupervised anomaly segmentation via deep feature reconstruction},
  author={Yong Shi and Jie Yang and Zhiquan Qi},
  journal={Neurocomputing},
  year={2021}
}

@inproceedings{MKD,
  title={Multiresolution knowledge distillation for anomaly detection},
  author={Salehi, Mohammadreza and Sadjadi, Niousha and Baselizadeh, Soroosh and Rohban, Mohammad H and Rabiee, Hamid R},
  booktitle={Proceedings of the IEEE/CVF conference on computer vision and pattern recognition},
  pages={14902--14912},
  year={2021}
}

@article{MBPFM,
  title={Unsupervised image anomaly detection and segmentation based on pre-trained feature mapping},
  author={Wan, Qian and Gao, Liang and Li, Xinyu and Wen, Long},
  journal={IEEE Transactions on Industrial Informatics},
  year={2022},
  publisher={IEEE}
}

@article{PaDiM,
  title={PaDiM: A Patch Distribution Modeling Framework for Anomaly Detection and Localization},
  author={Defard Thomas and Setkov Aleksandr and Loesch Angelique and Audigier Romaric},
  journal={Lecture Notes in Computer Science},
  year={2021}
}

@article{STMAE,
  title={Siamese Transition Masked Autoencoders as Uniform Unsupervised Visual Anomaly Detector},
  author={Yao, Haiming and Wang, Xue and Yu, Wenyong},
  journal={arXiv preprint arXiv:2211.00349},
  year={2022}
}

@article{Vit,
  title={An image is worth 16x16 words: Transformers for image recognition at scale},
  author={Dosovitskiy, Alexey and Beyer, Lucas and Kolesnikov, Alexander and Weissenborn, Dirk and Zhai, Xiaohua and Unterthiner, Thomas and Dehghani, Mostafa and Minderer, Matthias and Heigold, Georg and Gelly, Sylvain and others},
  journal={arXiv preprint arXiv:2010.11929},
  year={2020}
}

@inproceedings{PacthCore,
  title={Towards total recall in industrial anomaly detection},
  author={Roth, Karsten and Pemula, Latha and Zepeda, Joaquin and Sch{\"o}lkopf, Bernhard and Brox, Thomas and Gehler, Peter},
  booktitle={Proceedings of the IEEE/CVF Conference on Computer Vision and Pattern Recognition},
  pages={14318--14328},
  year={2022}
}

@ARTICLE{NDP-Net,
  author={Luo, Wei and Yao, Haiming and Yu, Wenyong},
  journal={IEEE Transactions on Instrumentation and Measurement}, 
  title={Normal Reference Attention and Defective Feature Perception Network for Surface Defect Detection}, 
  year={2023},
  volume={72},
  number={},
  pages={1-14},
  doi={10.1109/TIM.2023.3268658}}

@article{wideresnet,
  title={Wide residual networks},
  author={Zagoruyko, Sergey and Komodakis, Nikos},
  journal={arXiv preprint arXiv:1605.07146},
  year={2016}
}

@inproceedings{RD4AD,
  title={Anomaly detection via reverse distillation from one-class embedding},
  author={Deng, Hanqiu and Li, Xingyu},
  booktitle={Proceedings of the IEEE/CVF Conference on Computer Vision and Pattern Recognition},
  pages={9737--9746},
  year={2022}
}

@article{fastflow,
  title={Fastflow: Unsupervised anomaly detection and localization via 2d normalizing flows},
  author={Yu, Jiawei and Zheng, Ye and Wang, Xiang and Li, Wei and Wu, Yushuang and Zhao, Rui and Wu, Liwei},
  journal={arXiv preprint arXiv:2111.07677},
  year={2021}
}

@article{chen2022utrad,
  title={Utrad: Anomaly detection and localization with u-transformer},
  author={Chen, Liyang and You, Zhiyuan and Zhang, Nian and Xi, Juntong and Le, Xinyi},
  journal={Neural Networks},
  volume={147},
  pages={53--62},
  year={2022},
  publisher={Elsevier}
}

@article{salehi2020puzzle,
  title={Puzzle-ae: Novelty detection in images through solving puzzles},
  author={Salehi, Mohammadreza and Eftekhar, Ainaz and Sadjadi, Niousha and Rohban, Mohammad Hossein and Rabiee, Hamid R},
  journal={arXiv preprint arXiv:2008.12959},
  year={2020}
}

@article{cao2022informative,
  title={Informative knowledge distillation for image anomaly segmentation},
  author={Cao, Yunkang and Wan, Qian and Shen, Weiming and Gao, Liang},
  journal={Knowledge-Based Systems},
  volume={248},
  pages={108846},
  year={2022},
  publisher={Elsevier}
}

@inproceedings{mvgimagelevel,
  title={Modeling the distribution of normal data in pre-trained deep features for anomaly detection},
  author={Rippel, Oliver and Mertens, Patrick and Merhof, Dorit},
  booktitle={2020 25th International Conference on Pattern Recognition (ICPR)},
  pages={6726--6733},
  year={2021},
  organization={IEEE}
}

@article{guo2023mldfr,
  title={MLDFR: A Multilevel Features Restoration Method Based on Damaged Images for Anomaly Detection and Localization},
  author={Guo, Yinghui and Jiang, Meng and Huang, Qianhong and Cheng, Yang and Gong, Jun},
  journal={IEEE Transactions on Industrial Informatics},
  year={2023},
  publisher={IEEE}
}

@inproceedings{fod,
  title={Focus the Discrepancy: Intra-and Inter-Correlation Learning for Image Anomaly Detection},
  author={Yao, Xincheng and Li, Ruoqi and Qian, Zefeng and Luo, Yan and Zhang, Chongyang},
  booktitle={Proceedings of the IEEE/CVF International Conference on Computer Vision},
  pages={6803--6813},
  year={2023}
}

@inproceedings{liu2023simplenet,
  title={Simplenet: A simple network for image anomaly detection and localization},
  author={Liu, Zhikang and Zhou, Yiming and Xu, Yuansheng and Wang, Zilei},
  booktitle={Proceedings of the IEEE/CVF Conference on Computer Vision and Pattern Recognition},
  pages={20402--20411},
  year={2023}
}

@article{uniad,
  title={A unified model for multi-class anomaly detection},
  author={You, Zhiyuan and Cui, Lei and Shen, Yujun and Yang, Kai and Lu, Xin and Zheng, Yu and Le, Xinyi},
  journal={Advances in Neural Information Processing Systems},
  volume={35},
  pages={4571--4584},
  year={2022}
}

@inproceedings{RegAD,
  title={Registration based few-shot anomaly detection},
  author={Huang, Chaoqin and Guan, Haoyan and Jiang, Aofan and Zhang, Ya and Spratling, Michael and Wang, Yan-Feng},
  booktitle={European Conference on Computer Vision},
  pages={303--319},
  year={2022},
  organization={Springer}
}

@article{lu2023hierarchical,
  title={Hierarchical vector quantized transformer for multi-class unsupervised anomaly detection},
  author={Lu, Ruiying and Wu, YuJie and Tian, Long and Wang, Dongsheng and Chen, Bo and Liu, Xiyang and Hu, Ruimin},
  journal={arXiv preprint arXiv:2310.14228},
  year={2023}
}

@article{unified,
  title={A unified model for multi-class anomaly detection},
  author={You, Zhiyuan and Cui, Lei and Shen, Yujun and Yang, Kai and Lu, Xin and Zheng, Yu and Le, Xinyi},
  journal={Advances in Neural Information Processing Systems},
  volume={35},
  pages={4571--4584},
  year={2022}
}

@inproceedings{tdg,
  title={A hierarchical transformation-discriminating generative model for few shot anomaly detection},
  author={Sheynin, Shelly and Benaim, Sagie and Wolf, Lior},
  booktitle={Proceedings of the IEEE/CVF International Conference on Computer Vision},
  pages={8495--8504},
  year={2021}
}

@inproceedings{differnet,
  title={Same same but differnet: Semi-supervised defect detection with normalizing flows},
  author={Rudolph, Marco and Wandt, Bastian and Rosenhahn, Bodo},
  booktitle={Proceedings of the IEEE/CVF winter conference on applications of computer vision},
  pages={1907--1916},
  year={2021}
}

@article{AdamW,
  title={Decoupled weight decay regularization},
  author={Loshchilov, Ilya and Hutter, Frank},
  journal={arXiv preprint arXiv:1711.05101},
  year={2017}
}

@article{shen2025algorithm,
  title={An algorithm based on lightweight semantic features for ancient mural element object detection},
  author={Shen, Jiaquan and Liu, Ningzhong and Sun, Han and Li, Deguang and Zhang, Yongxin and Han, Lulu},
  journal={npj Heritage Science},
  volume={13},
  number={1},
  pages={70},
  year={2025},
  publisher={Springer International Publishing Cham}
}

@inproceedings{li2021cutpaste,
  title={Cutpaste: Self-supervised learning for anomaly detection and localization},
  author={Li, Chun-Liang and Sohn, Kihyuk and Yoon, Jinsung and Pfister, Tomas},
  booktitle={Proceedings of the IEEE/CVF conference on computer vision and pattern recognition},
  pages={9664--9674},
  year={2021}
}

@inproceedings{BTAD,
  title={VT-ADL: A vision transformer network for image anomaly detection and localization},
  author={Mishra, Pankaj and Verk, Riccardo and Fornasier, Daniele and Piciarelli, Claudio and Foresti, Gian Luca},
  booktitle={2021 IEEE 30th International Symposium on Industrial Electronics (ISIE)},
  pages={01--06},
  year={2021},
  organization={IEEE}
}

@inproceedings{he2024diffusion,
  title={A diffusion-based framework for multi-class anomaly detection},
  author={He, Haoyang and Zhang, Jiangning and Chen, Hongxu and Chen, Xuhai and Li, Zhishan and Chen, Xu and Wang, Yabiao and Wang, Chengjie and Xie, Lei},
  booktitle={Proceedings of the AAAI Conference on Artificial Intelligence},
  volume={38},
  number={8},
  pages={8472--8480},
  year={2024}
}

@inproceedings{fang2023fastrecon,
  title={Fastrecon: Few-shot industrial anomaly detection via fast feature reconstruction},
  author={Fang, Zheng and Wang, Xiaoyang and Li, Haocheng and Liu, Jiejie and Hu, Qiugui and Xiao, Jimin},
  booktitle={Proceedings of the IEEE/CVF International Conference on Computer Vision},
  pages={17481--17490},
  year={2023}
}

@inproceedings{hyun2024reconpatch,
  title={ReConPatch: Contrastive patch representation learning for industrial anomaly detection},
  author={Hyun, Jeeho and Kim, Sangyun and Jeon, Giyoung and Kim, Seung Hwan and Bae, Kyunghoon and Kang, Byung Jun},
  booktitle={Proceedings of the IEEE/CVF Winter Conference on Applications of Computer Vision},
  pages={2052--2061},
  year={2024}
}

@inproceedings{yao2023focus,
  title={Focus the discrepancy: Intra-and inter-correlation learning for image anomaly detection},
  author={Yao, Xincheng and Li, Ruoqi and Qian, Zefeng and Luo, Yan and Zhang, Chongyang},
  booktitle={Proceedings of the IEEE/CVF International Conference on Computer Vision},
  pages={6803--6813},
  year={2023}
}

@article{wang2025deep,
  title={Deep feature clustering for multi-class industrial image anomaly detection},
  author={Wang, Rongxiang and Li, Zhi and Zheng, Long and Wang, Weidong and Li, Shuyun},
  journal={Knowledge-Based Systems},
  pages={113134},
  year={2025},
  publisher={Elsevier}
}

@article{shen2021finger,
  title={Finger vein recognition algorithm based on lightweight deep convolutional neural network},
  author={Shen, Jiaquan and Liu, Ningzhong and Xu, Chenglu and Sun, Han and Xiao, Yushun and Li, Deguang and Zhang, Yongxin},
  journal={IEEE Transactions on Instrumentation and Measurement},
  volume={71},
  pages={1--13},
  year={2021},
  publisher={IEEE}
}

@article{wang2024mtdiff,
  title={MTDiff: Visual anomaly detection with multi-scale diffusion models},
  author={Wang, Xubin and Li, Wenju and He, Xiangjian},
  journal={Knowledge-Based Systems},
  volume={302},
  pages={112364},
  year={2024},
  publisher={Elsevier}
}

@article{bai2024dual,
  title={Dual-path frequency discriminators for few-shot anomaly detection},
  author={Bai, Yuhu and Zhang, Jiangning and Chen, Zhaofeng and Dong, Yuhang and Cao, Yunkang and Tian, Guanzhong},
  journal={Knowledge-Based Systems},
  volume={302},
  pages={112397},
  year={2024},
  publisher={Elsevier}
}

@article{jiang2023masked,
  title={A masked reverse knowledge distillation method incorporating global and local information for image anomaly detection},
  author={Jiang, Yuxin and Cao, Yunkang and Shen, Weiming},
  journal={Knowledge-Based Systems},
  volume={280},
  pages={110982},
  year={2023},
  publisher={Elsevier}
}

@inproceedings{hemambaad,
  title={MambaAD: Exploring State Space Models for Multi-class Unsupervised Anomaly Detection},
  author={He, Haoyang and Bai, Yuhu and Zhang, Jiangning and He, Qingdong and Chen, Hongxu and Gan, Zhenye and Wang, Chengjie and Li, Xiangtai and Tian, Guanzhong and Xie, Lei},
  booktitle={The Thirty-eighth Annual Conference on Neural Information Processing Systems}
}

@article{guo2024dinomaly,
  title={Dinomaly: The Less Is More Philosophy in Multi-Class Unsupervised Anomaly Detection},
  author={Guo, Jia and Lu, Shuai and Zhang, Weihang and Chen, Fang and Liao, Hongen and Li, Huiqi},
  journal={arXiv preprint arXiv:2405.14325},
  year={2024}
}

@article{mamba,
  title={Mamba: Linear-Time Sequence Modeling with Selective State Spaces},
  author={Gu, Albert and Dao, Tri},
  journal={arXiv preprint arXiv:2312.00752},
  year={2023}
}

@article{DINOV2-R,
  title={Vision transformers need registers},
  author={Darcet, Timoth{\'e}e and Oquab, Maxime and Mairal, Julien and Bojanowski, Piotr},
  journal={arXiv preprint arXiv:2309.16588},
  year={2023}
}

@InProceedings{Luo_2025_CVPR,
    author    = {Luo, Wei and Cao, Yunkang and Yao, Haiming and Zhang, Xiaotian and Lou, Jianan and Cheng, Yuqi and Shen, Weiming and Yu, Wenyong},
    title     = {Exploring Intrinsic Normal Prototypes within a Single Image for Universal Anomaly Detection},
    booktitle = {Proceedings of the IEEE/CVF Conference on Computer Vision and Pattern Recognition (CVPR)},
    month     = {June},
    year      = {2025},
    pages     = {9974-9983}
}

@article{shen2024instrument,
  title={An instrument indication acquisition algorithm based on lightweight deep convolutional neural network and hybrid attention fine-grained features},
  author={Shen, Jiaquan and Liu, Ningzhong and Sun, Han and Li, Deguang and Zhang, Yongxin},
  journal={IEEE Transactions on Instrumentation and Measurement},
  volume={73},
  pages={1--16},
  year={2024},
  publisher={IEEE}
}

@article{luo2024ami,
  title={AMI-Net: Adaptive mask inpainting network for industrial anomaly detection and localization},
  author={Luo, Wei and Yao, Haiming and Yu, Wenyong and Li, Zhengyong},
  journal={IEEE Transactions on Automation Science and Engineering},
  volume={22},
  pages={1591--1605},
  year={2024},
  publisher={IEEE}
}

\end{document}